\documentclass[10pt,twocolumn,letterpaper]{article}

\usepackage{iccv}
\usepackage{times}
\usepackage{epsfig}
\usepackage{graphicx}
\usepackage{amsmath}
\usepackage{amssymb}

\usepackage{multirow}
\usepackage{booktabs}
\usepackage{float}
\usepackage{subfigure}
\usepackage{bbding}

\usepackage{marvosym}

\newcommand\blfootnote[1]{%
  \begingroup
  \renewcommand\thefootnote{}\footnote{#1}%
  \addtocounter{footnote}{-1}%
  \endgroup
}

\usepackage[breaklinks=true,bookmarks=true, colorlinks]{hyperref}

\iccvfinalcopy 

\def\iccvPaperID{8674} 

\ificcvfinal\fi

\begin{document}

\title{Target Adaptive Context Aggregation for Video Scene Graph Generation}

\author{Yao Teng$^1$ \qquad Limin Wang\textsuperscript{$1$ \Letter} \qquad Zhifeng Li$^2$ \qquad Gangshan Wu$^1$\\
$^1$State Key Laboratory for Novel Software Technology, Nanjing University, China\\
$^2$Tencent AI Lab, Shenzhen, China\\
{\tt\small tengyao19980325@gmail.com, \{lmwang, gswu\}@nju.edu.cn, \tt\small michaelzfli@tencent.com}\\
}

\maketitle
\ificcvfinal\thispagestyle{empty}\fi

\begin{abstract}
   This paper deals with a challenging task of video scene graph generation (VidSGG), which could serve as a structured video representation for high-level understanding tasks. We present a new {\em detect-to-track} paradigm for this task by decoupling the context modeling for relation prediction from the complicated low-level entity tracking. Specifically, we design an efficient method for frame-level VidSGG, termed as {\em Target Adaptive Context Aggregation Network} (TRACE), with a focus on capturing spatio-temporal context information for relation recognition. Our TRACE framework streamlines the VidSGG pipeline with a modular design, and presents two unique blocks of Hierarchical Relation Tree (HRTree) construction and Target-adaptive Context Aggregation. More specific, our HRTree first provides an adpative structure for organizing possible relation candidates efficiently, and guides context aggregation module to effectively capture spatio-temporal structure information. Then, we obtain a contextualized feature representation for each relation candidate and build a classification head to recognize its relation category. Finally, we provide a simple temporal association strategy to track TRACE detected results to yield the video-level VidSGG. We perform experiments on two VidSGG benchmarks: ImageNet-VidVRD and Action Genome, and the results demonstrate that our TRACE achieves the state-of-the-art performance. The code and models are made available at \url{https://github.com/MCG-NJU/TRACE}.
\end{abstract}
\blfootnote{\Letter: Corresponding author.}

\section{Introduction}

\begin{figure}[t]
\begin{center}
\includegraphics[width=0.98\linewidth]{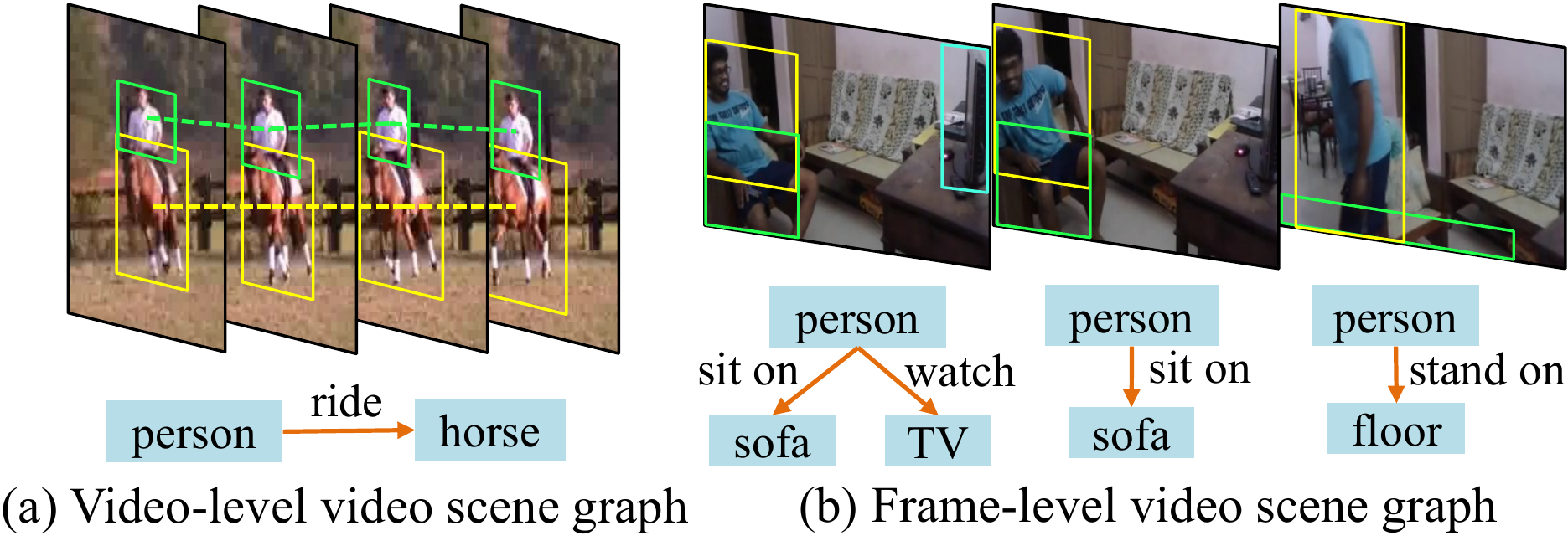}
\vspace{-2mm}
\end{center}
   \caption{(a) An example of video-level VidSGG. The subject/object nodes in this graph are entity trajectories and their relation is constant for this clip. (b) An example for frame-level VidSGG. The frame-level VidSGG is determined by short-term information for each frame and could vary along time.}
\label{Fig:example_introduce_vid}
\vspace{-4mm}
\end{figure}

Video understanding tasks, such as action recognition~\cite{two-stream,i3d,artnet,TSN}, temporal action localization~\cite{SSN,bmn,RTD}, spatio-temporal action detection~\cite{MOC,vidTransformer}, have received lots of research attention in the past few years. Most of these methods simply provide a single label or spatio-temporal extent of each action instance in a long video sequence. However, an ideal video understanding system is expected to not only recognize the action types, but also provide detailed and structured interpretation of the whole scene by parsing an event into a sequence of interactions between different visual entities.
This structured video representation, known as video scene graph~\cite{vidvrd}, can contribute to more accurate action recognition~\cite{ag} and allow for our vision models to tackle high-level and complex inference tasks, such as video caption~\cite{Youtube2text,VenugopalanRDMD15}, video retrieval~\cite{TALL,TVR}, and video question answering~\cite{MovieQA,TVQA}.
Nevertheless, video scene graph generation (VidSGG)~\cite{vidvrd-gcn, vidvrd-energy_graph, Beyond-Short-Term-Snippet, vidvrd-Multiple_Hypothesis_Association} has received much less research efforts in our community, when compared with image scene graph generation~\cite{Scene_graph_generation_by_iterative_message_passing,graphrcnn,F-net}.

As shown in Figure~\ref{Fig:example_introduce_vid}, the existing benchmark of VidSGG can be roughly grouped into two types according to the granularity of its graph representation: (1) video-level scene graph generation, where each graph node represents an object trajectory, and the edge captures the relation between visual entities, which is constant for one clip. (2) frame-level scene graph generation, where the graph is defined at frame level and the relation could change over time in this short clip. For video-level VidSGG, it requires to accurately trim long videos into short clips (e.g., 30 frames) in advance, according to the precise temporal boundaries of relations. This setting cannot be easily adapted to realistic VidSGG in untrimmed videos, as trimming is difficult and subjective due to temporal ambiguity. In contrast, frame-level VidSGG provides a more flexible mechanism for relation representation in continuous video streams. In addition, these frame-level VidSGG could easily yield the video-level scene graph by using temporal association to track adjacent results. However, previous works~\cite{vidvrd,vidvrd-gcn,Beyond-Short-Term-Snippet} on VidSGG mainly neglect the frame-level scene graphs, and directly recognize video-level relations based on the results of object tracking. As a result, they all yield a heavy pipeline highly dependent on tracking.

In this paper, we aim to present a new method to address the above two tasks simultaneously.
Our basic idea is to first generate the video scene graph at each frame by utilizing short-term video information, and then track each frame-level scene graph along time dimension to obtain the video-level result. We argue that this {\em detect-to-track} VidSGG paradigm will decouple the tasks of video relation recognition and temporal tracking, 
making our method focus more on modeling spatio-temporal context in videos.
The key of recognizing visual relation is that inferring the interaction between visual entities usually requires comprehensive understanding of spatio-temporal context information in the video.
For instance, recognizing whether a person is sitting on a sofa or standing from a sofa is based on the temporal variation of human movement with respect to sofa over time. 
Thus, we aim to devise a modular framework that can effectively determine and capture such complex spatio-temporal contextual information (e.g., temporal motion, object relation, person relation etc.) for efficient VidSGG.

Spatio-temporal context information is much more complex and diverse in videos than single images. To handle this issue, we design an efficient adaptive framework to select and propagate contextual information in videos, coined as TaRget Adaptive Context AggrEgation Network (TRACE). The key to TRACE is to organize relation candidates with an adaptive hierarchical relation tree (HRTree), and then perform target-adaptive context information aggregation for each relation candidate based on it. HRTree is not only helpful for the information aggregation, but also enables efficient processing of numerous relation candidates in a limited memory consumption. As for effective context information aggregation, we present an attentive module to fuse temporal information selectively and a directional propagation module to capture spatial structured information. Finally, the target-adaptive aggregated representation for each candidate can provide sufficient contextual information for relation classification. Furthermore, we employ a common temporal association algorithm to link frame-level graphs into a video-level result.

Specifically, we utilize a 3D CNN for extracting temporal features, a 2D CNN for extracting center frame representation, and an object detection network for object candidates with their visual features.
Based on these low-level visual representations, TRACE streamlines the VidSGG pipeline with modules of HRTree construction, context aggregation, relation classification, and optional temporal association. We evaluate TRACE on two datasets: Action Genome (AG)~\cite{ag} and ImageNet-VidVRD (VidVRD)~\cite{vidvrd}. AG is a brand-new dataset for frame-level VidSGG and only the methods for image SGG are evaluated on it, while VidVRD is a video-level VidSGG dataset. On AG, as a new specific method for frame-level VidSGG, TRACE achieves the state-of-the-art performance on the standard three evaluation modes: scene graph detection (SGDet), scene graph classification (SGCls), and predicate classification (PredCls). Specifically, TRACE outperforms the best model on average by 1.5\% and 1.2\% for mAP$_{rel}$~\cite{Graphical_Contrastive_Losses_for_Scene_Graph_Parsing} and mean Recalls~\cite{tang_treelstm} of the three modes, respectively, and get comparable performance at Recalls. On VidVRD, with a simple temporal linking strategy, our TRACE achieves good performance under the video-level metrics. Concretely, TRACE outperforms the best model with ground-truth trajectories and the same association algorithm by 2.8\%, 1.0\% and 2.1\% at mAP, Recall@50 and Recall@100 respectively. Given the same features, TRACE also outperforms the state-of-the-art model by 2.8\% at mAP. To sum up, our contributions are as follows:

\begin{figure*}
\begin{center}
\includegraphics[width=0.90\linewidth]{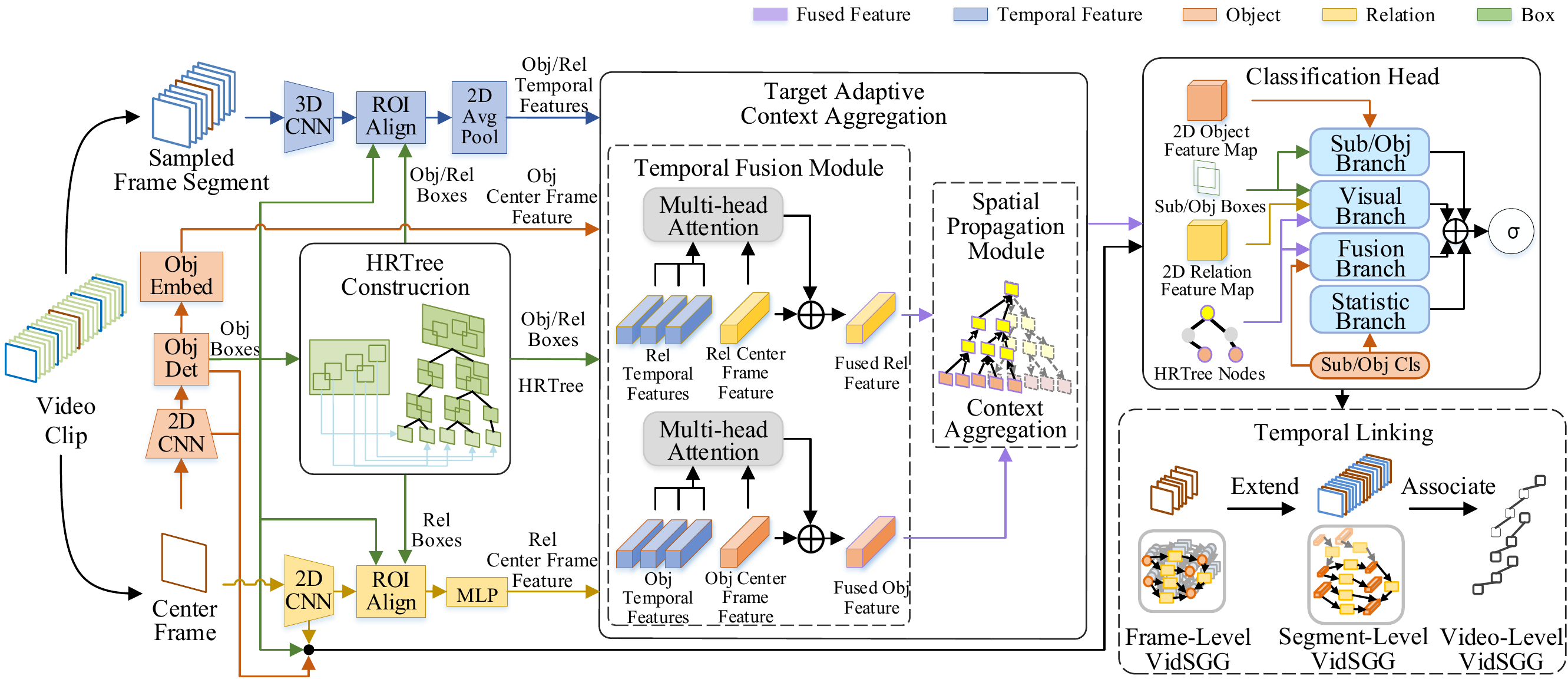}
\end{center}
\vspace{-4mm}
   \caption{\small {\bf TRACE framework.} Our TRACE framework is composed of feature extraction, hierarchical relation tree (HRTree) construction, context aggregation, classification head, and optional temporal linking, for VidSGG in a {\em detect-to-track} manner. Our model takes a clip as input and generates a spatio-temporal feature representation using a 3D CNN. The center frame is passed through an object detection network and a 2D CNN for static feature extraction. Based on these detection results, we build HRTree to organize relation candidates. The feature representation and HRTree are passed through a target-adaptive context aggregation module to obtain the contextualized representation for each relation candidate. This contextualized relation features is used to classify each candidate into relation classes. Finally, a simple temporal linking strategy is used to associate frame-level scene graphs to yield video-level results. The round-headed black arrow indicates transmitting the features directly. $\bigoplus$ denotes the operation of plus, and $\sigma$ denotes the sigmoid function. }
\label{Fig:overview}
\vspace{-3mm}
\end{figure*}

\begin{itemize}
  \item [1)]
  We propose a new {\em detect-to-track} paradigm for video-level VidSGG, coined as Target Adaptive Context Aggregation Network (TRACE). This new perspective decouples the context modeling for relation prediction from the complicated low-level entity tracking. With this new paradigm, we provide a baseline method for VidSGG and gains improvement over the state-of-the-art method on ImageNet-VidVRD dataset.
  \item [2)]
  As a pure frame-level VidSGG framework, TRACE presents a more modular framework to capture spatio-temporal context information for relation recognition than the previous methods, and obtains the best performance on Action Genome dataset.
  \item [3)]
  In our TRACE, we propose an adaptive structure called as hierarchical relation tree (HRTree). By using HRTree, the efficient context information aggregation among candidates is enabled. Moreover, our experiment demonstrates that this module allows us to save memory for more parameters, thus resulting a better performance than a fully connected graph.
\end{itemize}

\section{Related Work}
\noindent
\textbf{Scene Graph Generation (SGG).} Since the concept of scene graph was defined in \cite{image_retrieval_using_scene_graphs}, SGG task has become an important problem in computer vision. Along the research line, aggregating context information among interacted objects is quite effective for SGG. Xu et al.~\cite{Scene_graph_generation_by_iterative_message_passing} constructed a primal graph and utilize GRU~\cite{gru} to pass message between its nodes. Yang et al.~\cite{graphrcnn} utilized graph convolutional networks~\cite{gcn} and attention mechanism \cite{attention} for context information aggregation. Li et al. \cite{F-net} reduced the number of candidate relations by non-maximal suppression~\cite{nms}. Tang et al.~\cite{tang_treelstm}, Yin et al.~\cite{zoom-net} and Wang et al.~\cite{Sketching_Image_Gist} explored the application of tree structure in the domain of SGG from different aspects. In this paper, we argue that a better approach for graph construction is important for SGG frameworks. Therefore, we propose to combine the relation candidates hierarchically and adopt a tree structure for message passing between object and relation features directly, which is more memory efficient and differs from \cite{tang_treelstm,zoom-net,Sketching_Image_Gist}. 

\noindent
\textbf{Video Scene Graph Generation (VidSGG).} The concept of VidSGG was first proposed by Shang et al.~\cite{vidvrd} and they released a dataset named ImageNet-VidVRD. In \cite{vidvrd}, they used improved Dense Trajectories feature~\cite{idt} to predict the pairwise relations in video segments and then associated these relation triplets into video level. Subsequently, several works focused on video-level VidSGG have been released~\cite{vidvrd-gcn, vidvrd-energy_graph, Beyond-Short-Term-Snippet, vidvrd-Multiple_Hypothesis_Association}. 
However, all these methods used the {\em track-to-detect} paradigm and required complicated pre-processing to link detections into tubes. Thus, they heavily depended on the tracking results and lacked the flexibility of capturing relations in a frame level for more accurate results.
Recently, Ji et al.~\cite{ag} released Action Genome dataset which only focuses on frame-level VidSGG, which prompted us to consider unifying these tasks in a concise way.
In this paper, we present a new {\em detect-to-track} paradigm for VidSGG and could be used for both frame-level and video-level tasks of SGG. Our experiment results indicate that this new paradigm exhibit high flexibility and effectiveness for video-level VidSGG.

\section{Technical Approach}

\textbf{Overview.} As shown in Figure~\ref{Fig:overview}, we propose a new method for frame-level VidSGG, termed as {\em Target Adaptive Context Aggregation Network} (TRACE). The input of our model is a dense sampled short clip and its center frame. Our TRACE streamlines the VidSGG pipeline with the components of feature extraction, HRTree construction, context aggregation, relation classification, and optional temporal linking. {\em First}, objects are detected in the center frame. The spatial features in the center frame are extracted by using a 2D CNN and temporal features in the clip are extracted with a 3D CNN. Furthermore, the static object features are combined with word-embedding \cite{glove} for the subsequent blocks. {\em Second}, hierarchical relation tree (HRTree) is built to organize visual relation candidates in a compact and efficient way. {\em Third}, we perform target-adaptive context feature aggregation at a relatively low memory cost with the help of HRTree. Specifically, we devise a temporal attentive module for the fusion of temporal features. Then, a directional spatial aggregation module is responsible for propagating context information. {\em Finally}, a classification module is used for inferring the relation class of each relation candidate. Additionally, our method can be extended to the video level with a simple temporal association strategy.

\subsection{Hierarchical Relation Tree Construction}
\label{subsec:hrtree}

Hierarchical Relation Tree (HRTree) for relation candidates organization is built in a hierarchical bottom-up way. The leaf nodes in HRTree represent the objects detected in the center frame. The non-leaf nodes are derived from their child nodes and represent their composite relations. Specifically, HRTree is constructed in a progressive manner based on spatial proximity. Given the spatial coordinates of nodes in one layer, we use Gaussian kernel function to calculate the sum of pairwise similarity for each node:
\vspace{-1.5mm}
\begin{equation}
\begin{gathered}
\mathrm{score}_k = \sum_{i}{ e^{-\Vert f_k - f_i\Vert^2}},
\vspace{-2.5mm}
\end{gathered}
\end{equation}
where score$_k$ encodes the relative location information for node $k$ and $f$ represents the spatial coordinates. After obtaining the scores of nodes in one layer, we sort the nodes based on their scores and select part of them as the centers. Then, the other nodes are merged into the centers closest to them measured by their spatial union. Therefore, the updated centers form the parent layer of the current layer and this process is repeated until there is one node left. Euclidean distance is used as a measure of distance. 

\begin{figure}[t]
\centering
\subfigure[Scheme 1.]{
\begin{minipage}[t]{0.44\linewidth}
\centering
\includegraphics[width=0.4\linewidth]{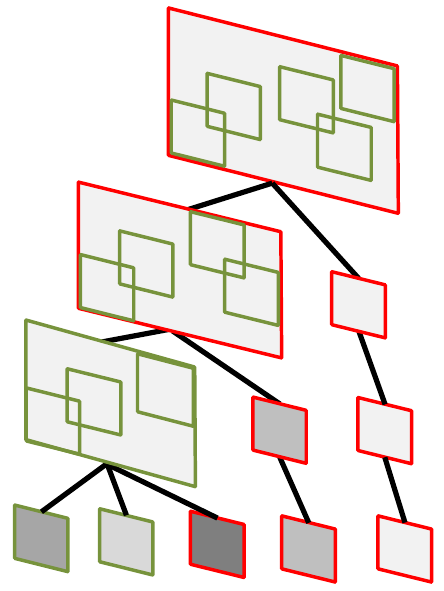}
\end{minipage}
} 
\subfigure[Scheme 2.]{
\begin{minipage}[t]{0.44\linewidth}
\centering
\includegraphics[width=0.4\linewidth]{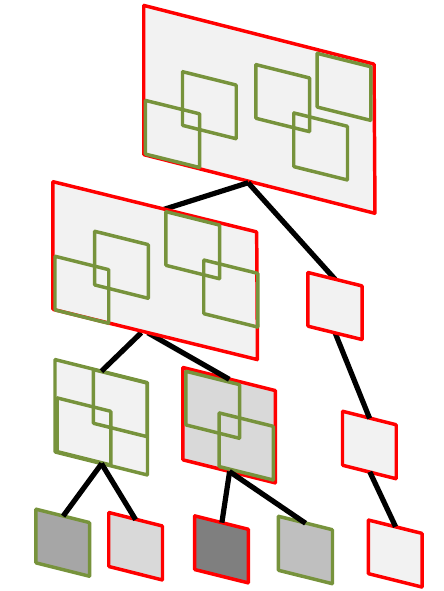}
\end{minipage}
}
\centering
   \vspace{1mm}
   \caption{The two schemes for HRTree construction. The shades of color represent the scores and the darker color means the higher score. The edges of centers are in red.}
\label{Fig:hrtree}
\vspace{-2mm}
\end{figure}

With regard to the selection of centers in each layer, as shown in Figure \ref{Fig:hrtree}, we propose two implementation schemes: (1) From the highest to the lowest score, one node is chosen as the center every other node. (2) We fix the number of nodes to be rounded half of the total number of their child nodes. Then, we select half of the nodes from the part with the highest score and the others from the part with the lowest score as the centers. In our case, the number of visual relation candidates is $O(n) $, which means relation candidates greatly decrease in quantity compared with previous fully-connected graph, thus saving more computational and memory overhead for context aggregation.

\subsection{Target Adaptive Context Aggregation}
\textbf{Temporal Fusion Module.} After introducing the construction of HRTree, we are ready to describe how use this structure to guide context information aggregation. First, we describe the temporal context information fusion in this subsection and then the spatial context aggregation in next subsection. As analyzed above, temporal motion information is important to recognize some relations such as {\em wiping}. So, we follow the common practice of extracting temporal features with a box tube~\cite{vidTransformer}.
We use a 3D CNN to extract spatio-temporal features to provide motion information for relation candidates.

\begin{figure}[t]
\begin{center}
\includegraphics[width=0.87\linewidth]{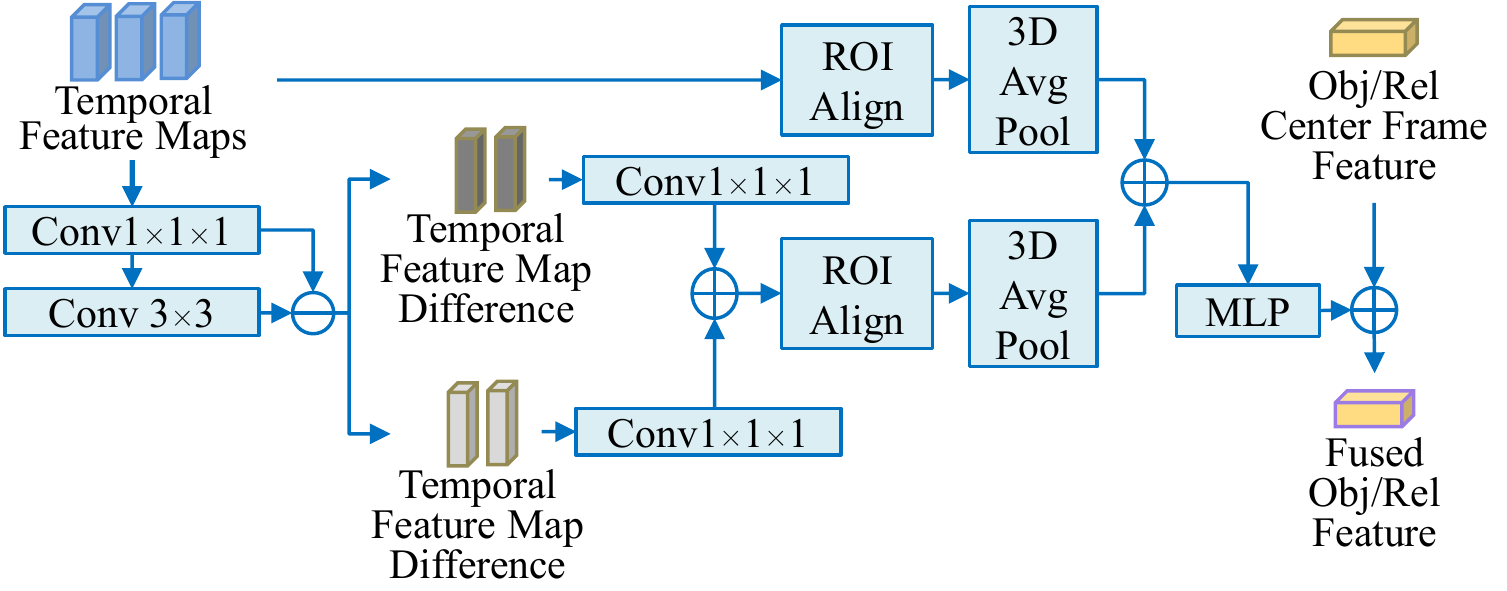}
\end{center}
   \vspace{-4.5mm}
   \caption{The temporal difference module for temporal fusion.}
\label{Fig:temporal2}
\vspace{-3mm}
\end{figure}

Specifically, for each non-leaf node, i.e., relation candidate, we extract a feature representation corresponding to this relation candidate from 3D CNN feature map. It is implemented by first stretching the candidate bounding box along time with repetition to form a tube. Then, we extract a feature at each time point with the corresponding box in the tube by using the standard RoI Align operation~\cite{roialign}. The resulting features across time are used for temporal information aggregation for current relation candidate.  We propose two ways of fusing temporal information: (1) As shown in Figure \ref{Fig:overview}, the multi-head attention mechanism~\cite{transformer} is applied to these temporal features with the spatial feature as query. It is essentially the weighted sum of 3D features and the weights are learnt adaptively based on 2D features. (2) As shown in Figure \ref{Fig:temporal2}, a temporal difference operation is applied to the output of 3D backbone to extract motion features, and simple average pooling operation is employed for temporal fusion. In the experiment section (see Sec. \ref{subsec:ablation}), we show that these two kinds of temporal fusion modules are effective for the certain types of relations related to motion (e.g., {\em writing on}, {\em carrying}). However, for some short-term relation recognition, its improvement is not so evident.

\textbf{Spatial Propagation Module.} In this subsection, we describe the spatial context aggregation mechanism based on HRTree. Specifically, we adopt a group tree-GRU scheme for context aggregation in bidirectional progation manner. The features of nodes in HRTree are divided into multiple groups across the feature dimension. Then, features in each group of are fed into an independent tree-GRU~\cite{Modelling_Sentence_Pairs_with_Tree-structured_Attentive_Encoder}. In each tree-GRU, bottom-up feature aggregation is performed at first. Then, the top-down feature refinement which is equivalent to a common GRU~\cite{gru} takes place. Subsequently, a multi-layer perceptron (MLP) is applied for the concatenation of features to yield the contextualized features. In experiment, we observe that this spatial propagation module is effective to aggregate spatial context information for relation recognition.

\begin{figure}[t]
\begin{center}
\includegraphics[width=0.88\linewidth]{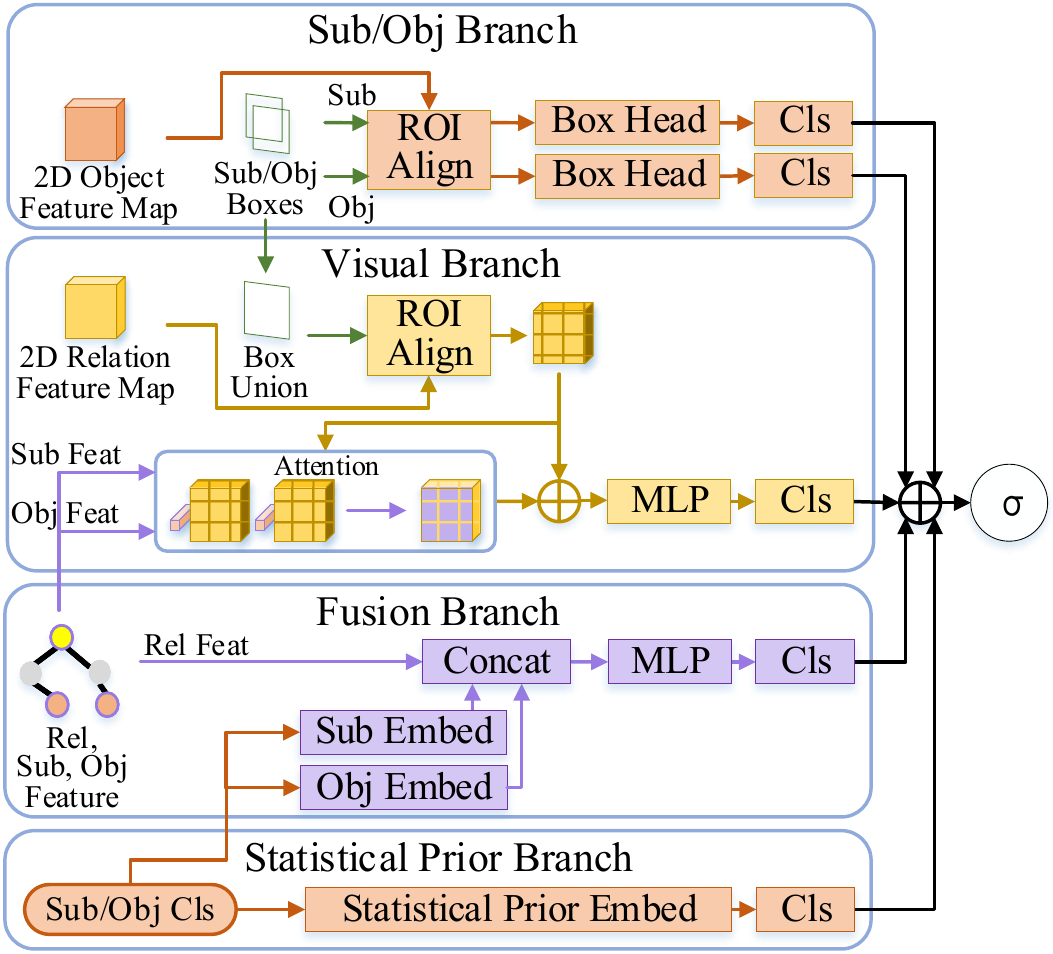}
\end{center}
   \vspace{-4mm}
   \caption{Illustration of the classification head.}
\label{Fig:cls}
\vspace{-2mm}
\end{figure}

\subsection{Classification Head}
As shown in Figure \ref{Fig:cls}, the classification head is responsible for relation inference. It consists of four branches and each branch provides one result. The final score for classification is the sum of them followed by a sigmoid function.

\textbf{Visual Branch.} A relation feature map is generated by applying ROI Align~\cite{roialign} to the 2D CNN backbone's output with the union of pairwise object proposals. After obtaining each relation feature map, the feature vectors of the subject and object perform an attention mechanism on its dimension-reduced version. Specifically, the score map is first generated by the cosine similarity of the feature map and the feature vector on each pixel, and then the attention map is derived from the element-wise product between the feature map and the post-softmax score map. The attention maps and feature maps are used for classification.

\textbf{Fusion Branch.} We first input subject and object classification scores into word-embedding blocks~\cite{glove}. Then, the embedded feature vectors are concatenated with the contextualized relation feature from the spatial propagation module for a classifier. The contextualized relation feature corresponding to the concatenated vector of subject and object belongs to their least common ancestors in HRTree.

\textbf{Subject/Object Branch and Statistical Prior Branch.} We adopt the same subject/object structure as~\cite{Graphical_Contrastive_Losses_for_Scene_Graph_Parsing}.
Following~\cite{Graphical_Contrastive_Losses_for_Scene_Graph_Parsing,neural_motifs}, we also employ the subject/object classification statistics as our input for better results.

\subsection{Temporal Linking}
Finally, we describe the temporal linking strategy to fuse frame-level scene graphs to the video-level results.
The long video clip is first divided into overlapping video segments (e.g., 30 frames for one segment, 15 frames for the interval) and then the tracking is performed on each segment. The object trajectories obtained from tracking are used for this linking.
As for one video segment, we sample a quarter of the frames with frame-level scene graphs for linking. If one triplet appears in only one frame, it is directly counted with its predicted score. For triplets with the same predicted categories in multiple frames, the triplet is counted once with summed scores if their subjects and objects belong to the same trajectory, respectively.
As for the whole video, the triplets among two neighbouring segments are associated only if their predicted categories are the same and their viou of subject/object trajectories is beyond a threshold of 0.5~\cite{vidvrd}.
The video-level scores can be either average~\cite{vidvrd} or maximum~\cite{vidvrd-gcn}.
In a greedy manner, high-scoring triplets take precedence over other ones during the association process.

\section{Experiments}
In this section, we present experimental results on two datasets: ImageNet-VidVRD~\cite{vidvrd} and Action Genome~\cite{ag}. First, we report the evaluation settings and implementation details. Then, we show the ablation studies and comparisons to state-of-the-art methods.

\subsection{Evaluation Settings}
\textbf{ImageNet-VidVRD (VidVRD).} VidVRD~\cite{vidvrd} focuses on a wider range of relations not limited to human-object interaction. Especially, the subjects in VidVRD could be various categories more than just humans. VidVRD contains 35 object categories and 132 relation categories. Unlike the traditional SGG datasets, such as VG~\cite{vg}, multiple relations could occur between a subject and an object in VidVRD~\cite{vidvrd}. The annotations of VidVRD are spatio-temporal originally, so the label of one relation contains both spatial location and duration. Different from~\cite{vidvrd, vidvrd-gcn}, in our work, we assign the spatio-temporal annotations to each frame. After the conversion, the average number of relations and objects is 9.7 and 2.5 in each frame respectively. Furthermore, the number of relations in each object pair is about 2.0 on average.
In the same way as~\cite{vidvrd, vidvrd-gcn}, Recalls and mAP of relation detection are used to evaluate our model. Relation Tagging~\cite{vidvrd} is also considered for comparison. Furthermore, consistent with \cite{vidvrd}, top 20 predicted relations for each pair of objects are kept for evaluation. The threshold for viewing a predicted box as a hit is 0.5.

\textbf{Action Genome (AG).} AG~\cite{ag} is a dataset bridging human action and human-object relations. In AG, the relations are all human action and all subjects belong the category of {\em person}. AG is built on Charades~\cite{charades} dataset so it contains a large number of indoor scenes.
The number of object and relation categories in AG is 36 and 26 respectively. Similar to VidVRD~\cite{vidvrd}, multiple relations may exist between subjects and objects in AG. After preprocessing, the number of relations and objects is 7.3 and 3.2 in each frame on average respectively. Furthermore, the number of relations in each object pair is 3.3 on average. Furthermore, the triplets with overlapped subject bounding box and object bounding box in AG is more than 85\%.
In line with~\cite{ag}, we adopt three modes for the evaluation on AG: scene graph detection (SGDet), scene graph classification (SGCls) and predicate classification (PredCls)~\cite{visual_relationship_detection_with_language_priors}. The traditional specific metric for the three standard modes is Recall. 
However, due to the imbalanced distribution of relations, we introduce mean Recall (mR)~\cite{tang_treelstm}, mAP$_{rel}$ and wmAP$_{rel}$~\cite{Graphical_Contrastive_Losses_for_Scene_Graph_Parsing} to AG.
The predicted boxes that have an IoU of at least 0.5 with the ground-truth boxes are counted as a hit.
It is worth noting that since the relations in AG are all human-object interaction,
PredCls and SGCls in our benchmark provide not only the ground-truth object boxes but also the potentially related pairs.
Due to the multiple relations, the graph constraint which restricts each pair of objects to one prediction of triplet is not suitable here. 
Additionally, to avoid the situation where the predictions randomly hit the ground-truth triplets, each pair of objects is only allowed corresponding to $k$ predictions and $k$ is set to 6 or 7.

\begin{table*}
\small
\begin{center}
\setlength{\tabcolsep}{2pt}
\begin{tabular}{c|c|c|c|c|c|c|c|c}
\hline
Temp Fusion & beneath & carrying & eating & lying on  & standing on & touching & wiping & writing on  \\ 
\hline
\hline
- & 47.65   & 12.51   & 19.98   & 19.55    & 45.40   & 35.39   & 4.71   & 25.63  \\ \hline
1 & 46.70(-0.95) & 14.96(+2.45) & 20.61(+0.63) & 26.64(+7.09)  & 44.54(-0.86) & 36.76(+1.37) & 6.27(+1.56) & 26.71(+1.08)  \\ \hline
2 & 47.7(+0.05)  & 13.74(+1.23) & 19.89(-0.09) & 19.10(-0.45)  & 46.08(+0.68) & 35.79(+0.40) & 5.10(+0.39) & 27.08(+1.45)  \\ 
\hline
\end{tabular}
\end{center}
\vspace{-2mm}
\caption{The Recall@20 (\%) of partial relation categories on AG \cite{ag} with top 6 Predictions for each pair. The format of values except the first line is the model's output with different temporal fusion scheme and its difference compared to the first line.}
\label{Tab:temporal_sh}
\vspace{-1mm}
\end{table*}

\begin{table*}
\small
\parbox{.29\linewidth}{
\centering
\setlength{\tabcolsep}{2pt}
\begin{tabular}{c|ccc}
\hline
 Fusion & mAP    & R@50  & R@100 \\
\hline
\hline
FC-G & 28.11 & 16.96 & 21.78 \\
\hline
- & 28.62 & 17.48 & 23.56 \\
Tree-2 & 29.28  & 17.70 & 22.65 \\
Tree-1 & \textbf{29.32}  & \textbf{18.45} & \textbf{23.85} \\
\hline
\end{tabular}
\vspace{1mm}
\caption{\textbf{Study on the context fusion structure}. We compare the models with different context aggregation methods.}
\label{Tab:ablation_tree}
}
\hfill
\parbox{.29\linewidth}{
\centering
\setlength{\tabcolsep}{2pt}
\begin{tabular}{c|ccc}
\hline
 Fusion & mAP    & R@50  & R@100 \\
\hline
\hline
- & \textbf{29.94} & 18.01 & 23.56 \\
Temp-2 & 29.80 & 18.08 & 23.23 \\
Temp-1 & 29.32  & \textbf{18.45} & \textbf{23.85} \\
\hline
\end{tabular}
\vspace{4.68mm}
\caption{\textbf{Study on the temporal fusion structure}. We compare the models with different temporal fusion modules.}
\label{Tab:ablation_temporal}
}
\hfill
\parbox{.35\linewidth}{
\centering
\setlength{\tabcolsep}{2pt}
\begin{tabular}{c|cccccc}
\hline
Fusion & R@20 & R@50 & mR@20 & mR@50 \\
\hline\hline
FC-G$^\ast$    & 32.32 & 44.63  & \textbf{27.60} & 38.17 \\
Tree-1$^\ast$   & 32.24 & 44.60 & 26.94 & 37.60 \\
\hline
Tree-1 & \textbf{33.41}  & \textbf{45.67} & 27.58 & \textbf{38.61} \\
\hline
\end{tabular}
\vspace{1mm}
\caption{We compare the tree structure in our model with ResNet-50-FPN~\cite{fpn} when the parameters are reduced at SGDet on AG. $^\ast$ means the parameters of the model are reduced.}
\label{Tab:ablation_tree_reduce}
}

\vspace{1.5mm}

\parbox{.3\linewidth}{
\centering
\setlength{\tabcolsep}{2pt}
\begin{tabular}{c|ccc}
\hline
Group Number & mAP    & R@50  & R@100 \\
\hline
\hline
2 & \textbf{29.81} & \textbf{18.51} & 23.72 \\
4 & 29.32  & 18.45 & \textbf{23.85} \\
\hline
\end{tabular}
\vspace{4.95mm}
\caption{\textbf{Study on the group number}. We compare TRACE with different group number in the context aggregation structure.}
\label{Tab:ablation_group}
}
\hfill
\parbox{.32\linewidth}{
\centering
\setlength{\tabcolsep}{2pt}
\begin{tabular}{c|ccc}
\hline
Model & mAP    & R@50  & R@100 \\
\hline
\hline
VidVRD-C~\cite{vidvrd} & 7.17 & 4.36 & 5.36 \\
Liu's~\cite{Beyond-Short-Term-Snippet}(Obj) & 14.01 & \textbf{8.47} & \textbf{11.00} \\
\textbf{Ours} & \textbf{15.06} & 7.67 & 10.32 \\
\hline
\end{tabular}
\vspace{5.6mm}
\caption{\textbf{Study on the framework}. We compare TRACE to other models with only object features.} 
\label{Tab:ablation_framework}
}
\hfill
\parbox{.34\linewidth}{
\centering
\setlength{\tabcolsep}{2pt}
\begin{tabular}{c|ccc}
\hline
Model & R@20  & R@50  & R@100 \\
\hline
\hline
RelDN~\cite{Graphical_Contrastive_Losses_for_Scene_Graph_Parsing} & 23.95 & 35.39 & 42.91 \\
\textbf{Ours} & \textbf{24.80} & \textbf{36.52} & \textbf{45.33} \\
\hline
\end{tabular}
\vspace{1mm}
\caption{\textbf{Study on the context aggregation structure without temporal fusion}. We compare our model without temporal fusion to RelDN~\cite{Graphical_Contrastive_Losses_for_Scene_Graph_Parsing} under the frame-level metrics on VidVRD~\cite{vidvrd}.} 
\label{Tab:ablation_pure_hrtree}
}
\vspace{-3mm}
\end{table*}

\subsection{Implementation Details}
The input of our model is a video segment sampled from the video clip and its center frame. The segment except the center frame is composed of $T=8 $ neighboring frames of the center frame with a temporal stride $v=4 $.

\textbf{Loss.} The loss is the weighted sum of the binary cross entropy for relations and cross entropy for the objects. The weight for relations is 1.0 while the weight for objects is 0.05. The relations are predicted by the classification module while the objects participating in the context aggregation and subsequent blocks are predicted by a classifier which is a duplicate of the one in Faster R-CNN~\cite{fasterrcnn}. Notably, when testing, this classifier is not activated.

\textbf{Training.} We use RTX 2080ti with 11G GPU memory for training. In line with \cite{ag, vidvrd}, Faster R-CNN~\cite{fasterrcnn} with ResNet~\cite{res} pretrained on COCO \cite{Microsoft_coco_Common_objects_in_context} is first trained on each dataset.
We utilize 2D ResNet-50~\cite{res} to extract relation feature on the center frame, and use I3D ResNet-50~\cite{i3d} pretrained on Kinetics \cite{The_kinetics_human_action_video_dataset} to extract temporal information.
All layers in the backbone for object feature extraction are frozen when training TRACE.
We use SGD with momentum to optimize TRACE with batch size 1.
The initial learning rate for AG and VidVRD is set to 0.01 and 0.025 respectively.
The ratio between the foreground relations and the background relations is 1:3, and 2048 relations with 512 objects are used for training.
A triplet is defined as foreground if its relation and object classes are identical to that of a ground truth and its objects have an overlap of $iou > 0.5$ with that of the ground-truth respectively. The other triplets are background. We randomly select at most m foreground and k background relations. We set m as 512 and ensure m + k = 2048.
In particular, for VidVRD, due to the conversion from video segments to frames, we randomly select 15.9\% frames from the training set to train TRACE.

\textbf{Testing.} Top 100 object proposals are kept after object detection and per-class non-maximal suppression~\cite{nms} with an IoU of 0.5 is used in each frame. Due to the most objects in AG touching each other, following~\cite{Graphical_Contrastive_Losses_for_Scene_Graph_Parsing}, we only predict the relations in pairs with overlapped bounding boxes for SGDet. However, this trick is not applied in VidVRD.

\subsection{Ablation Studies}
\label{subsec:ablation}
We carry out ablation studies on VidVRD dataset. Recalls and mAP are adopted for evaluations. Aside from these, a per category breakdown experiment in Table~\ref{Tab:temporal_sh} illustrates the effectiveness of our temporal fusion module.

\textbf{Study on the context aggregation structure.} We begin our ablation study by exploring the effectiveness of the context aggregation structure in TRACE. 
We implement fairly comparable models by removing this module and changing the scheme of HRTree. However, due to the large memory cost of the fully connected graph (FC-G) which forms a complete bipartite graph with its pairwise relation nodes, we reduce the parameters of FC-G and report its performance. 
In Table~\ref{Tab:ablation_tree}, TRACE with HRTree of each scheme outperforms the model without fusion under the three metrics. Due to the reduction of parameters, the performance of FC-G is quite lower than the others in this table.
TRACE with HRTree of scheme 1 consumes 10.5GB and 6.6GB GPU memory for training and testing per batch, respectively.
However, we found that our model with FC-G and complete parameters consumes too memory to train, so we reduce the parameters to the same level for fair comparison between FC-G and HRTree. Table~\ref{Tab:ablation_tree_reduce} shows the results of FC-G is quite comparable to HRTree and the number of parameters is important. It illustrates that HRTree reduces the memory cost without much performance decrease.
Furthermore, we compare our model without temporal fusion module to RelDN~\cite{Graphical_Contrastive_Losses_for_Scene_Graph_Parsing} in Table~\ref{Tab:ablation_pure_hrtree}. It demonstrates the effectiveness of our pure frame-level context aggregation.

\textbf{Study on the temporal fusion module.} We compare the different schemes of temporal fusion module. Moreover, We report the result of TRACE without the temporal fusion. In Table~\ref{Tab:ablation_temporal}, The performance of TRACE with temporal fusion module of scheme 1 at Recalls is better than the model with temporal fusion of scheme 2 and the model without temporal fusion, but it is worse at mAP. For further research, we conduct a per category breakdown experiment. As shown in Table~\ref{Tab:temporal_sh}, the temporal fusion scheme 1 drastically improves the performance at servel relation categories such as {\em writing on}, {\em lying on} and {\em carrying}. However, the attention across temporal dimension for each relation candidates may disturb the spatial information, which leads to decreases in {\em beneath} and {\em standing on}. Scheme 2 does not adopt the adaptive temporal information aggregation and the increases on each category are not salient.

\textbf{Study on the {\em detect-to-track} framework.} We conduct experiments to show the effectiveness of our {\em detect-to-track} framework. Since almost all previous works are in the manner of {\em track-then-detect}~\cite{vidvrd,vidvrd-gcn,Beyond-Short-Term-Snippet}, we compare TRACE to VidVRD-C~\cite{vidvrd} and Liu's~\cite{Beyond-Short-Term-Snippet} with only object features. In Table~\ref{Tab:ablation_framework}, mAP of our model without the fusion is better than Liu's~\cite{Beyond-Short-Term-Snippet} while the Recalls are comparable.
Thus, the performance of our framework is comparable to that of the {\em track-then-detect} models, but more flexible.

\textbf{Study on the selection for group number.} The operation of grouping can reduce the computational complexity. In Table~\ref{Tab:ablation_group}, we find that the group number only affect the performance at mAP on VidVRD evidently. Moreover, during training, we found that the one group version consumed too memory to run.

\begin{table}[t]
\small
\begin{center}
\setlength{\tabcolsep}{2pt}
\begin{tabular}{ccccccc}
\hline
\multirow{2}{*}{Method} & \multicolumn{3}{c}{Relation Detection} & \multicolumn{3}{c}{Relation Tagging} \\ \cmidrule(lr){2-4} \cmidrule(lr){5-7}
 & mAP & R@50 & R@100 & P@1 & P@5 & P@10 \\ \hline\hline
VidVRD gt~\cite{vidvrd} & 15.53 & 12.51 & 16.55 & 43.50 & 29.70 & 23.20\\
VRD-GCN gt~\cite{vidvrd-gcn} & 26.52 & 17.50 & 21.80 & 62.50 & 44.20 & 31.10\\
\textbf{Ours gt} & \textbf{29.32} & \textbf{18.45} & \textbf{23.85} & \textbf{65.50} & \textbf{45.60} & \textbf{33.75} \\ \hline
VidVRD$^\dagger$~\cite{vidvrd} & 8.58 & 5.54 & 6.37 & 43.00 & 28.90 & 20.80 \\
GSTEG~\cite{vidvrd-energy_graph} & 9.52 & 7.05 & 7.67 & 51.50 & 39.50 & 28.23 \\
MHRA~\cite{Multiple_Hypothesis_Video_Relation_Detection} & 13.27 & 6.82 & 7.39 & 41.00 & 28.70 & 20.95 \\
VRD-GCN$^\dagger$~\cite{vidvrd-gcn} & 14.23 & 7.43 & 8.75 & \textbf{59.50} & 40.50 & 27.85\\
\textbf{Ours}$^\dagger$ & \textbf{15.81} & \textbf{8.07} & \textbf{10.30} & 56.00 & \textbf{44.50} & \textbf{32.95} \\
\hline
Liu's~\cite{Beyond-Short-Term-Snippet} & 14.81 & \textbf{9.14} & \textbf{11.39} & 55.50 & 38.90 & 28.90 \\
\textbf{Ours$^\ddagger$} & \textbf{17.57} & 9.08 & 11.15 & \textbf{61.00} & \textbf{45.30} & \textbf{33.50} \\
\hline
\end{tabular}
\end{center}
\vspace{-2mm}
\caption{The metrics~\cite{vidvrd} (\%) of various models on VidVRD. For fair comparison, we compare our method with~\cite{Beyond-Short-Term-Snippet} by using the object features and I3D features. $^\dagger$ denotes using the basic temporal linking proposed in~\cite{vidvrd} with average scores, while $^\ddagger$ means using maximal scores.}
\label{Tab:vidvrd}
\vspace{-2.7mm}
\end{table}

\begin{table*}
\small
\begin{center}
\setlength{\tabcolsep}{2pt}
\begin{tabular}{cccccccccccccc}
\hline
\multirow{3}{*}{\begin{tabular}[c]{@{}c@{}}Top k Predictions \\ for Each Pair\end{tabular}} & \multirow{3}{*}{Method} & \multicolumn{4}{c}{PredCls} & \multicolumn{4}{c}{SGCls} & \multicolumn{4}{c}{SGDet} \\ \cmidrule(lr){3-6} \cmidrule(lr){7-10} \cmidrule(lr){11-14}
& & \multicolumn{2}{c}{image}  & \multicolumn{2}{c}{video} & \multicolumn{2}{c}{image} & \multicolumn{2}{c}{video} & \multicolumn{2}{c}{image} & \multicolumn{2}{c}{video} \\ \cmidrule(lr){3-4} \cmidrule(lr){5-6} \cmidrule(lr){7-8} \cmidrule(lr){9-10} \cmidrule(lr){11-12} \cmidrule(lr){13-14}
 & & R@20   & R@50 & R@20 & R@50 & R@20 & R@50 & R@20 & R@50 & R@20 & R@50 & R@20 & R@50 \\
 \hline\hline
\multirow{4}{*}{k=7} & Freq Prior \cite{neural_motifs}   & 87.95 & 93.02 & 86.01 & 88.59 & 45.10 & 48.87 & 44.47 & 46.39 & 34.41 & 44.34 & 32.50 & 41.11 \\
       & G-RCNN \cite{graphrcnn}   & 88.73 & 93.73 & 86.28 & 88.93 & 45.57 & 49.75 & 45.11 & 47.22 & 34.28 & 44.47 & 32.60 & 41.29 \\
       & RelDN \cite{Graphical_Contrastive_Losses_for_Scene_Graph_Parsing}      & 90.89 & 96.09 & 88.77 & 91.43 & 46.47 & 50.31 & 45.87 & 47.78 & 34.92 & 45.27 & 33.18 & 42.10 \\
       & \textbf{Ours}  & \textbf{91.60} & \textbf{96.35} & \textbf{89.31} & \textbf{91.72} & \textbf{46.66} & \textbf{50.46} & \textbf{46.03} & \textbf{47.92} & \textbf{35.09} & \textbf{45.34} & \textbf{33.38} & \textbf{42.18} \\
 \hline
\multirow{4}{*}{k=6} & Freq Prior \cite{neural_motifs}   & 85.89 & 89.43 & 83.33 & 84.99 & 44.90 & 47.15 & 43.57 & 44.63 & 34.47 & 43.69 & 32.38 & 40.24 \\
       & G-RCNN \cite{graphrcnn}   & 87.03 & 90.60 & 84.02 & 85.74 & 45.82 & 48.31 & 44.60 & 45.77 & 34.60 & 43.98 & 32.75 & 40.65 \\
       & RelDN \cite{Graphical_Contrastive_Losses_for_Scene_Graph_Parsing}      & 89.63 & 93.56 & 87.01 & 88.86 & 46.76 & 49.11 & 45.48 & 46.57 & 35.22 & 44.94 & 33.39 & 41.64 \\
       & \textbf{Ours}  & \textbf{90.34} & \textbf{93.94} & \textbf{87.56} & \textbf{89.24} & \textbf{47.00} & \textbf{49.32} & \textbf{45.71} & \textbf{46.79} & \textbf{35.41} & \textbf{45.06} & \textbf{33.59} & \textbf{41.76} \\
 \hline
\end{tabular}
\end{center}
\vspace{-2mm}
\caption{Recall (\%) of various models with ResNet-101~\cite{res} on AG. For fair comparison, we reproduce the methods based on our object detection and the same training strategy, and our model performs better than the others.}
\vspace{-2mm}
\label{Tab:ag_recall}
\end{table*}

\begin{table}
\small
\begin{center}
\setlength{\tabcolsep}{0.1pt}
\begin{tabular}{ccccccc}
\hline
\multirow{3}{*}{Method} & \multicolumn{2}{c}{PredCls} & \multicolumn{2}{c}{SGCls} & \multicolumn{2}{c}{SGDet} \\ \cmidrule(lr){2-3} \cmidrule(lr){4-5} \cmidrule(lr){6-7}
 & mR@20  & mR@50 & mR@20  & mR@50 & mR@20  & mR@50 \\ 
\hline\hline
Freq Prior \cite{neural_motifs}    & 55.17   & 63.67    & 34.30   & 36.96  & 24.89   & 34.07 \\ 
G-RCNN \cite{graphrcnn}  & 56.32 & 61.31  & 36.19 & 38.29  & 27.79 & 34.99 \\
RelDN \cite{Graphical_Contrastive_Losses_for_Scene_Graph_Parsing}   & 59.81   & 63.47  &  39.92   & 41.93    & 30.39   & 39.53  \\
\textbf{Ours}   & \textbf{61.80} & \textbf{65.37}  & \textbf{41.19} & \textbf{43.21}  & \textbf{30.84} & \textbf{40.12} \\
\hline
\end{tabular}
\end{center}
\vspace{-2mm}
\caption{Mean recall~\cite{tang_treelstm} (\%) of various models with ResNet-101~\cite{res} on all images in AG. The number of triplets per frame is set to a limit of 50 and top 6 predictions for each pair are kept when evaluating.}
\label{Tab:ag_mrecall}
\vspace{-2mm}
\end{table}

\begin{table}
\small
\begin{center}
\setlength{\tabcolsep}{0.1pt}
\begin{tabular}{ccccccc}
\hline
\multirow{3}{*}{Method} & \multicolumn{2}{c}{PredCls} & \multicolumn{2}{c}{SGCls} & \multicolumn{2}{c}{SGDet} \\ \cmidrule(lr){2-3} \cmidrule(lr){4-5} \cmidrule(lr){6-7}
 & $\mathrm{mAP}_{r}$  & $\mathrm{wmAP}_{r}$ & $\mathrm{mAP}_{r}$  & $\mathrm{wmAP}_{r}$ & $\mathrm{mAP}_{r}$  & $\mathrm{wmAP}_{r}$ \\ 
 \hline\hline
Freq Prior \cite{neural_motifs}  & 33.10   & 65.92 & 14.29   & 22.68 & 9.45 & 15.58 \\ 
G-RCNN \cite{graphrcnn} & 41.21 & 70.89 & 17.64 & 22.53 & 11.76 & 15.90 \\
RelDN \cite{Graphical_Contrastive_Losses_for_Scene_Graph_Parsing}  & 50.08   & 72.26  & 20.07   & 23.88  & 12.93   & 15.94   \\
\textbf{Ours}  & \textbf{53.27} & \textbf{75.45} & \textbf{20.71} & \textbf{24.61} & \textbf{13.43} & \textbf{16.56} \\
\hline
\end{tabular}
\end{center}
\vspace{-2mm}
\caption{$\mathrm{mAP}_{rel}$ and $\mathrm{wmAP}_{rel}$ (\%) of various models with ResNet-101~\cite{res} on all images in AG. The number of triplets per frame is set to a limit of 50 and top 6 predictions for each pair are kept when evaluating. $\mathrm{mAP}_{r}$ and $\mathrm{wmAP}_{r}$ indicate $\mathrm{mAP}_{rel}$ and $\mathrm{wmAP}_{rel}$, respectively.}
\vspace{-2mm}
\label{Tab:ag_map}
\end{table}

\begin{figure}[t]
\begin{center}
\includegraphics[width=0.99\linewidth]{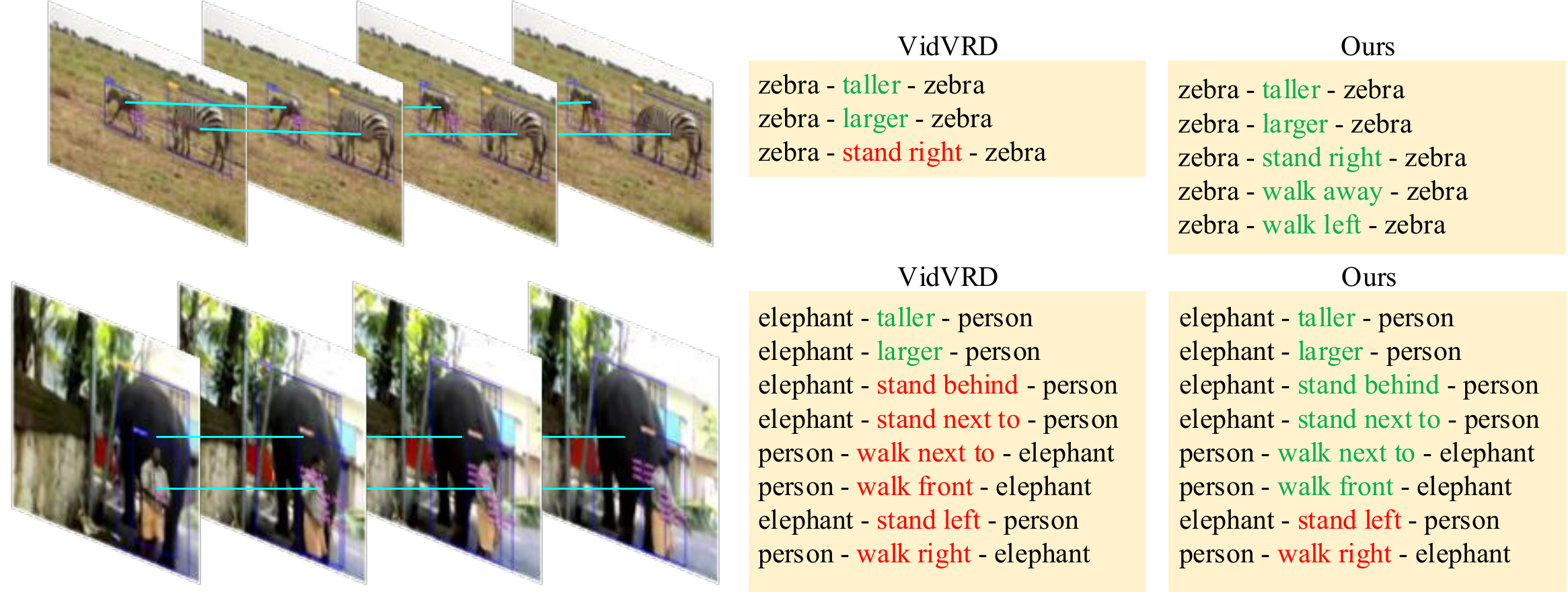}
\end{center}
   \vspace{-3mm}
   \caption{The Visualization of predictions hitting the ground-truth from VidVRD and TRACE. The correct results are marked in green. The relations with incomplete duration are marked in red. }
\label{Fig:vidvrd_vis}
\vspace{-2.5mm}
\end{figure}

\subsection{Comparison with the State of the Art}
\textbf{ImageNet-VidVRD (VidVRD).} As shown in Table \ref{Tab:vidvrd}, different conditions lead to different best performing models on VidVRD. With the ground-truth trajectories and the same association algorithm, our model outperforms the best method, VRD-GCN~\cite{vidvrd-gcn}, by 2.8\%, 1.0\% and 2.1\% at mAP, Recall@50 and Recall@100 respectively. Furthermore, under the condition of using trajectories provided by~\cite{vidvrd} and the basic association~\cite{vidvrd}, we push the performance at mAP, Recall@50, Recall@100 to 15.8\%, 8.1\% and 10.3\% respectively. Furthermore, like the siamese association in VRD-GCN~\cite{vidvrd-gcn}, we modify the calculation of scores in the basic association algorithm from averaging to maximizing and get improvements under all metrics. Compared to the state-of-the-art method, Liu's~\cite{Beyond-Short-Term-Snippet}, with both object features and I3D features, TRACE is comparable to it at Recalls, and beyond 2.8\% at mAP. Following~\cite{vidvrd,vidvrd-gcn}, we also report the results on relation tagging~\cite{vidvrd} and TRACE achieves good performance under different conditions.

\textbf{Action Genome (AG).} The results are summarized in Table~\ref{Tab:ag_recall}, Table~\ref{Tab:ag_mrecall} and Table~\ref{Tab:ag_map}. However, in Table~\ref{Tab:ag_recall}, the difference between the performance of various models at SGDet is not obvious. We analyze that the results at SGDet heavily depends on the object detector while the objects labeled in AG is far fewer than those in standard object detection datasets. Meanwhile, for fair comparison, we use the same detector for all methods and thus the difference is quite small. The performance at Recalls of SGDet in this dataset tends to be saturate. Therefore, PredCls and SGCls are more significant than SGDet in AG. Moreover, mAP$_{rel}$, wmAP$_{rel}$ and mR are more balanced metrics than Recalls, and are better metrics for revealing the gap between the performance of methods. Specifically, in Table \ref{Tab:ag_mrecall} and Table \ref{Tab:ag_map}, TRACE outperforms RelDN \cite{Graphical_Contrastive_Losses_for_Scene_Graph_Parsing} by 3.2\% and 0.6\% at mAP$_{rel}$ of PredCls and SGCls, by 2.0\% and 1.3\% at mRs of PredCls and SGCls on average.

\subsection{Qualitative Results}
Our qualitative results are shown in Figure~\ref{Fig:vidvrd_vis}. VidVRD detects few relation triplets in some scenes and fails to detect enough relations containing motion information, such as {\em walk away} and {\em walk left}. 
We analyze that its temporal fusion structure fails to find fine-grained changes of objects in the scenes with slow motion compared to ours. 
Moreover, the relations detected in VidVRD lasts for shorter duration than TRACE. 
It illustrates that our {\em detect-to-track} framework performs better than the {\em track-then-detect} one, due to the less disturbance of noisy tracking results.

\section{Conclusion}
In this paper we have proposed a modular framework, coined as Target Adaptive Context Aggregation Network (TRACE) for frame-level VidSGG. To adaptively and efficiently capture spatio-temporal context information, we design a new hierarchical relation tree to guide temporal attentive fusion and spatial message propagation. 
Our method combined with a simple temporal association strategy yields a modular video-level VidSGG baseline, obtaining the best performance without using complex tracking features under video-level metrics on ImageNet-VidVRD. 
For pure frame-level VidSGG task, TRACE still achieves new state-of-the-art results on benchmarks of Action Genome.

\paragraph{\bf Acknowledgements.} \small{This work is supported by National Natural Science Foundation of China (No. 62076119, No. 61921006), Program for Innovative Talents and Entrepreneur in Jiangsu Province, and Collaborative Innovation Center of Novel Software Technology and Industrialization, Tencent AI Lab Rhino-Bird Focused Research Program (No. JR202025)}.

\appendix

\section{Quantitative Analysis}

The partial results on mean Recalls~\cite{tang_treelstm} of various methods are shown in Table~\ref{Tab:ag_mrecall2}.

\begin{table}[h]
\small
\begin{center}
\setlength{\tabcolsep}{0.1pt}
\begin{tabular}{ccccccc}
\hline
\multirow{3}{*}{Method} & \multicolumn{2}{c}{PredCls} & \multicolumn{2}{c}{SGCls} & \multicolumn{2}{c}{SGDet} \\ \cmidrule(lr){2-3} \cmidrule(lr){4-5} \cmidrule(lr){6-7}
 & mR@20  & mR@50 & mR@20  & mR@50 & mR@20  & mR@50 \\ 
 \hline\hline
Freq Prior~\cite{neural_motifs} & 58.79   & 70.50   & 36.21   & 40.28    & 25.46   & 35.86   \\ 
G-RCNN~\cite{graphrcnn} & 59.61 & 67.39  & 37.80 & 41.43  & 28.61 & 37.06 \\
RelDN~\cite{Graphical_Contrastive_Losses_for_Scene_Graph_Parsing} & 71.27   & 80.68     & 41.79   & 45.23    & 30.97   & 41.42  \\
\textbf{Ours}    & \textbf{73.60} & \textbf{82.67}  & \textbf{42.69} & \textbf{46.32}  & \textbf{31.31} & \textbf{41.82}  \\
\hline
\end{tabular}
\end{center}
\vspace{-3mm}
\caption{Mean recall~\cite{tang_treelstm} (\%) of various models with ResNet-101~\cite{res} on all images in AG. The number of triplets per frame is set to a limit of 50 and top 7 predictions for each pair are kept when evaluating.}
\label{Tab:ag_mrecall2}
\vspace{-2mm}
\end{table}

\section{Qualitative Analysis}

\subsection{Visualization of the per-class performance} 
\begin{figure}[h]
\begin{center}
\includegraphics[width=0.98\linewidth]{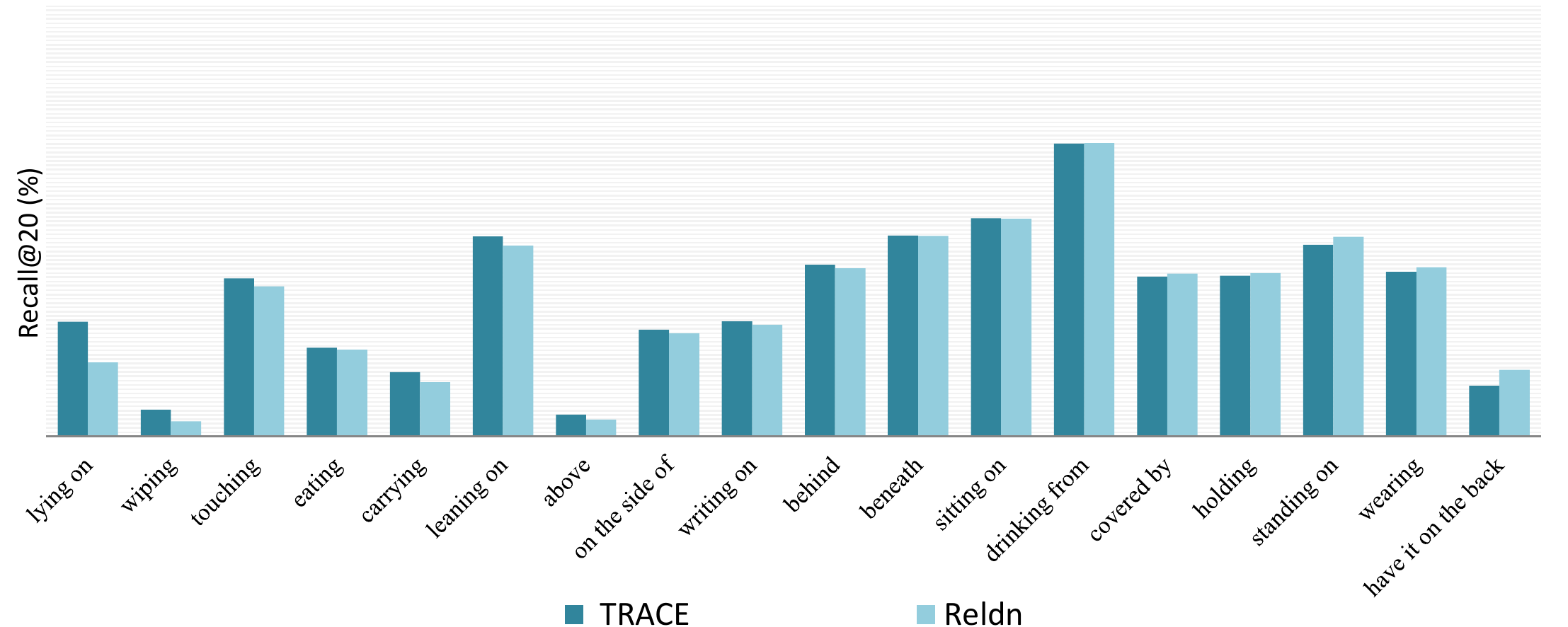}
\end{center}
   \caption{The per-class performance of our model compared to the state-of-the-art.}
\label{Fig:comp_reldn_temporal}
\vspace{-1mm}
\end{figure}

As shown in Figure~\ref{Fig:comp_reldn_temporal}, we compare our method to RelDN with per-class analysis for frame-level VidSGG. Our method performs well on motion-related classes such as {\em lying on}, {\em wiping} etc.

\subsection{Visualization of 2D relation feature} 
\begin{figure*}
\begin{center}
\includegraphics[width=0.7\linewidth]{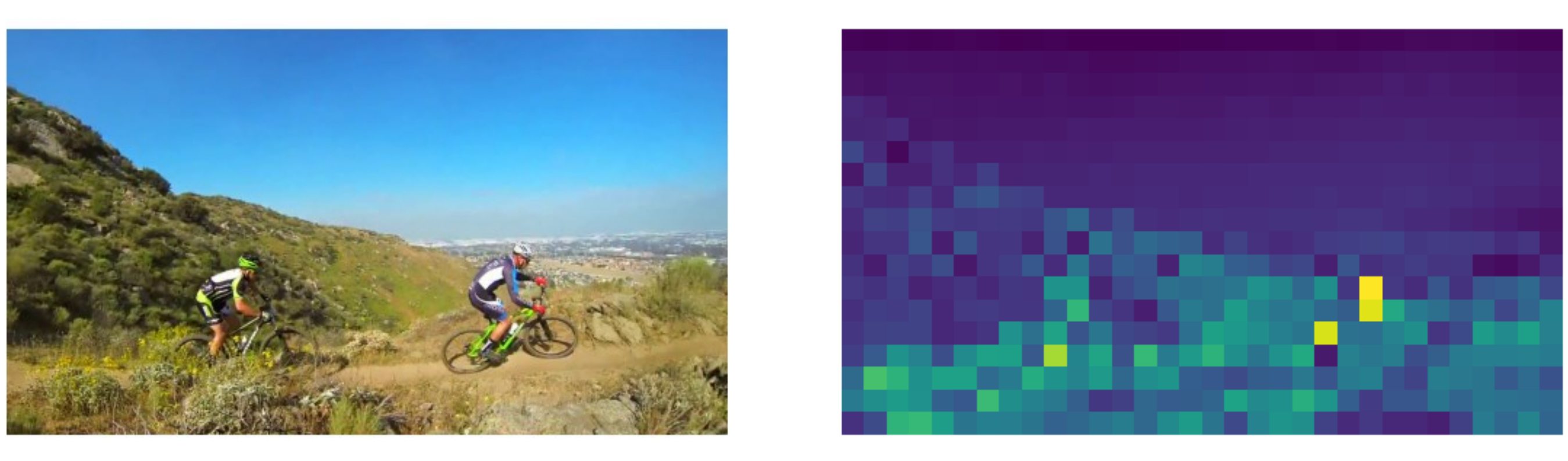}
\end{center}
   \caption{The visualization of the frame and the corresponding feature map produced by 2D ResNet-50~\cite{res}.}
\label{Fig:2dfeat}
\vspace{-2.5mm}
\end{figure*}

In Figure~\ref{Fig:2dfeat}, we provide the origin image and its feature map produced by 2D ResNet-50~\cite{res}. The highlighted areas represent the high activation. In this figure, the activation of the man with the bicycle is far greater than that of the background, which illustrates that our model obtains the direct interaction between the visual entities.

\subsection{Visualization of 3D relation feature} 
\begin{figure*}
\begin{center}
\includegraphics[width=0.7\linewidth]{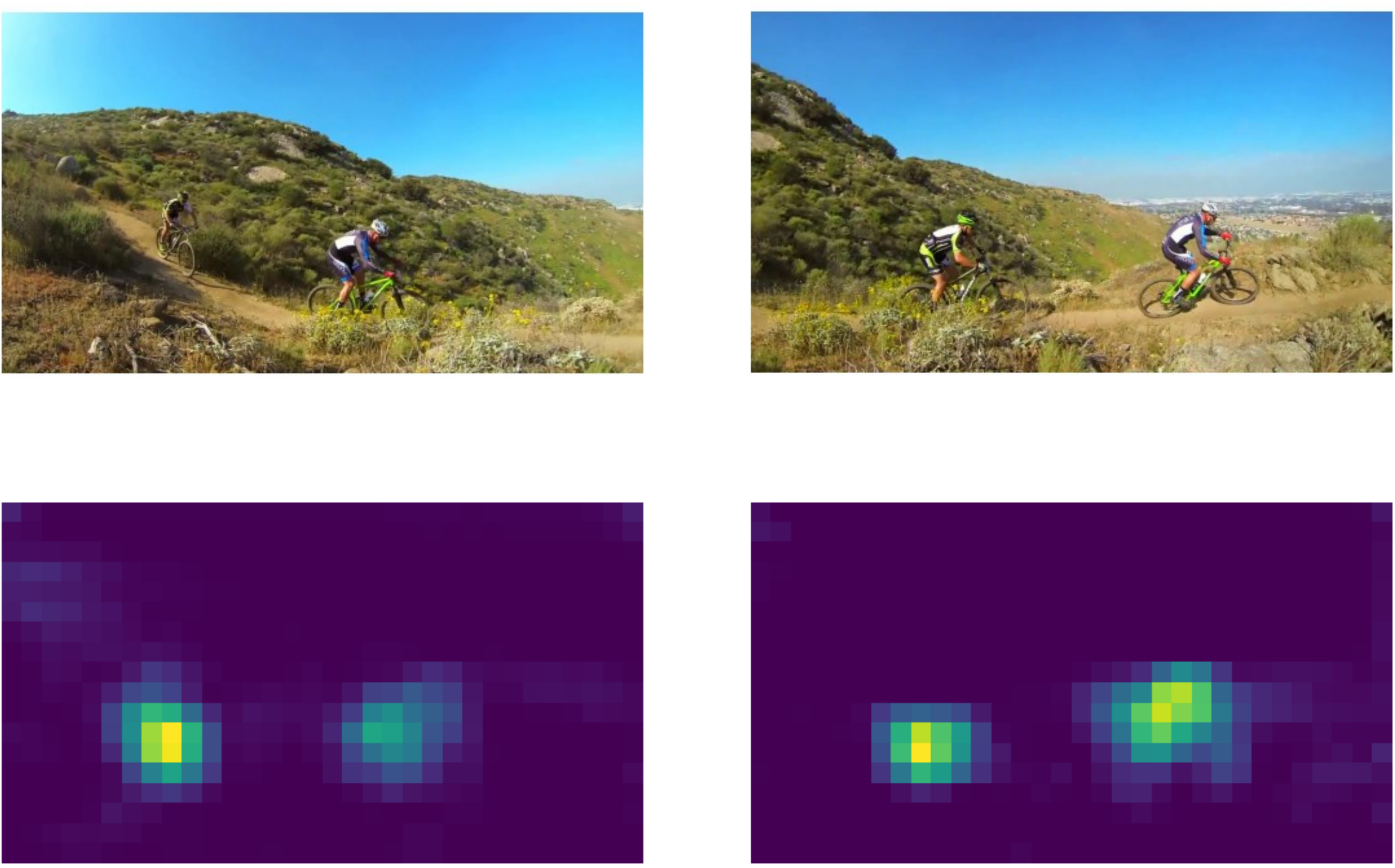}
\end{center}
   \caption{The visualization of the frames and the corresponding spatio-temporal feature map. The center frame and another frame ahead of it in temporal dimension are presented on the top line. The corresponding feature maps produced by an I3D ResNet-50~\cite{i3d} of the short video clip are presented on the bottom line across the temporal dimension.}
\label{Fig:3dfeat}
\vspace{-2.5mm}
\end{figure*}

In Figure~\ref{Fig:3dfeat}, we provide the feature maps produced by an I3D ResNet-50~\cite{i3d} of the short video clip whose center frame is the image mentioned above. We also provide another frame ahead of it in temporal dimension. In this figure, the bicycle with the man is the entities with the most motion information. The I3D appropriately obtains the movement changes in the video clip and the high activation in the feature map from the I3D indicates the movement.

\subsection{Visualization of Hierarchical Relation Tree} 
\begin{figure*}
\begin{center}
\includegraphics[width=0.5\linewidth]{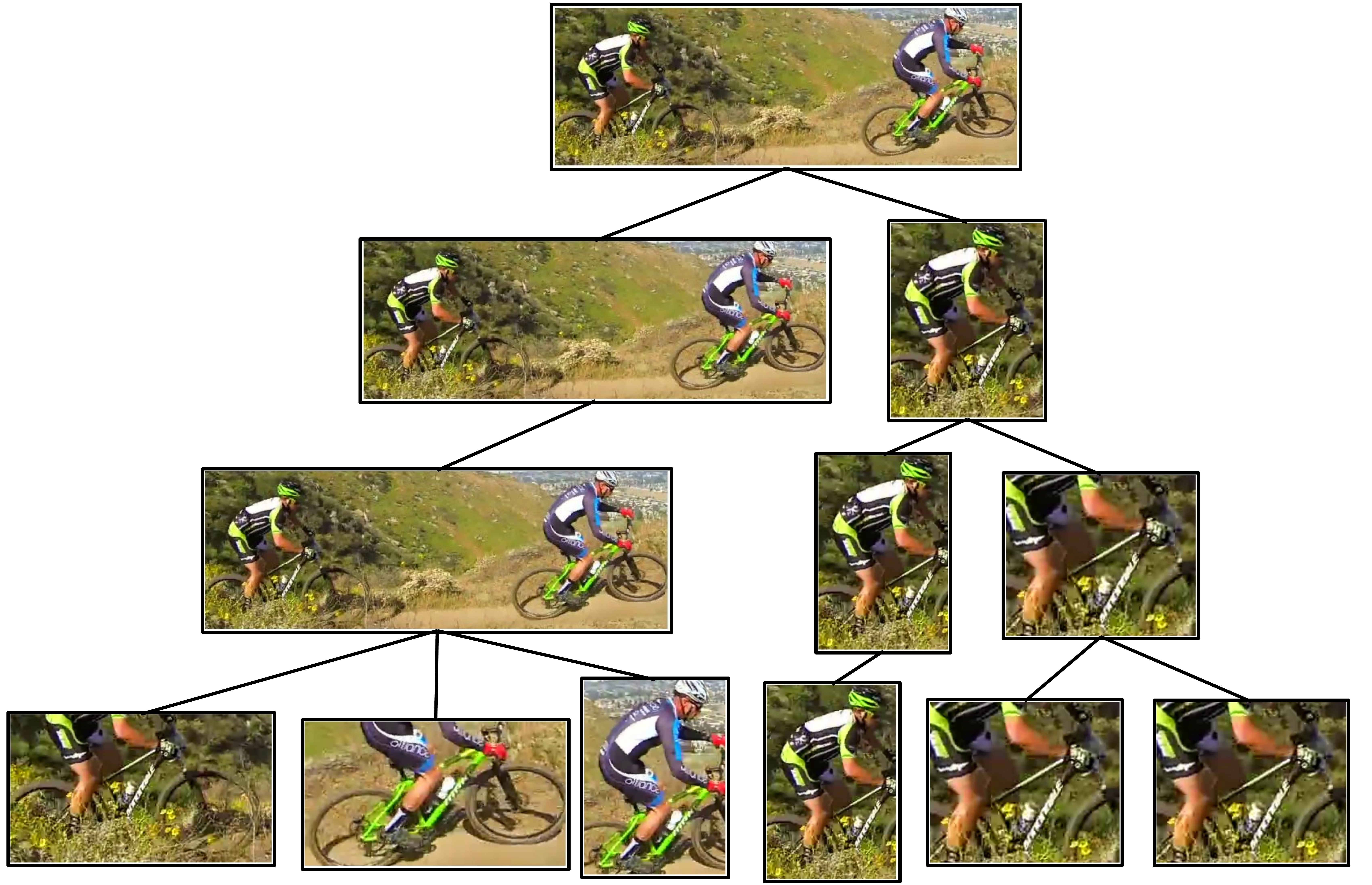}
\end{center}
   \caption{The visualization of Hierarchical Relation Tree. Each node is filled by the feature generated from the corresponding patch in the image. The repeated leaf nodes are attributed to the object detection algorithm.}
\label{Fig:hrtree_vis}
\vspace{-2.5mm}
\end{figure*}

In Figure~\ref{Fig:hrtree_vis}, we provide our Hierarchical Relation Tree in one frame. In this figure, the tree is built in a bottom-up manner and gradually expands the scope of its spatial coverage. Therefore, at the lower levels of the tree, our framework has the potential to obtain fine-grained relation feature, while at the top it obtains coarse-grained and long-distance relation feature.

\subsection{Visualization of the results}
\begin{figure*}
\centering

\begin{minipage}[t]{0.47\linewidth}
\centering
\includegraphics[width=1.0\linewidth]{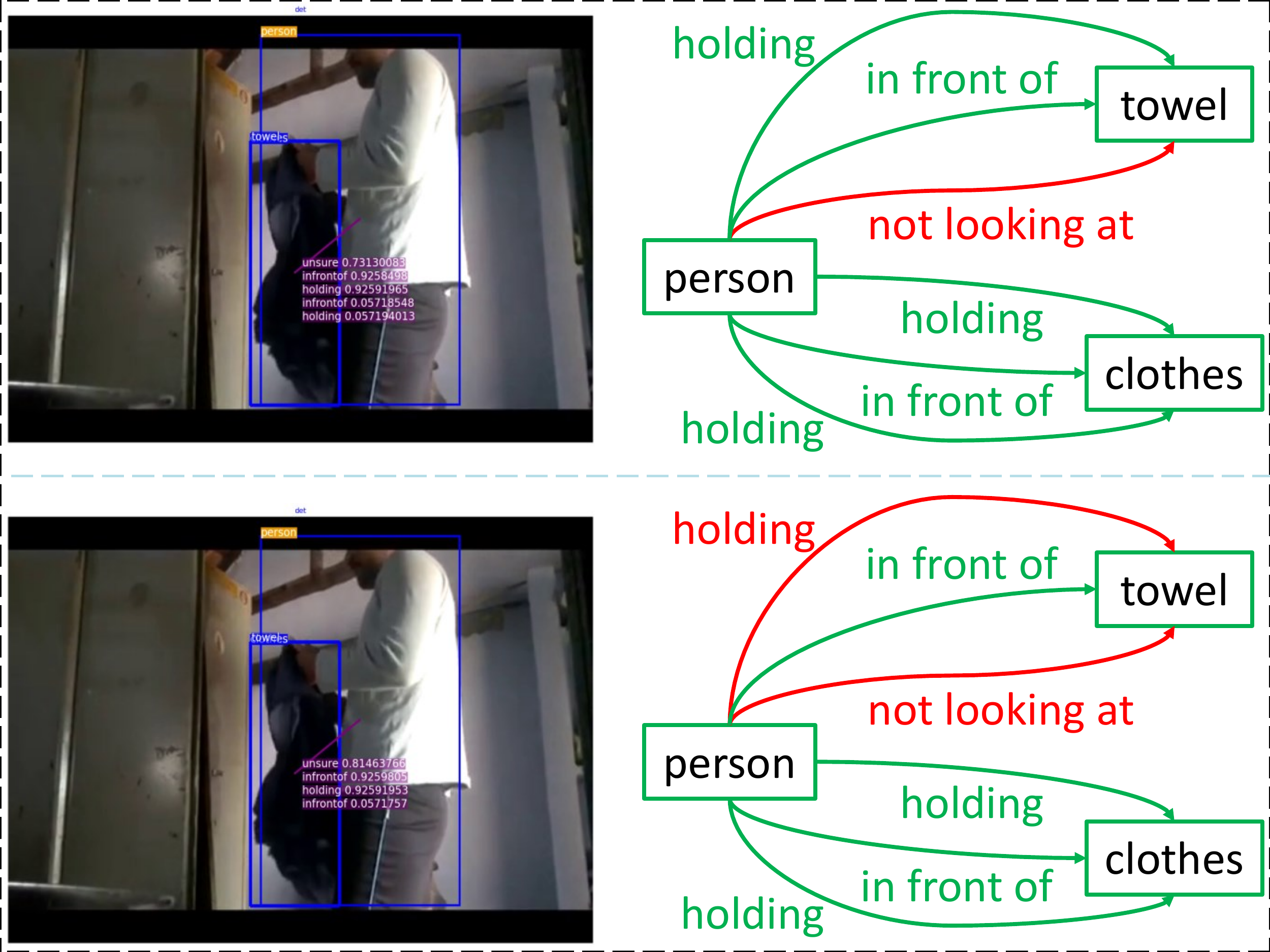}
\end{minipage}
\begin{minipage}[t]{0.47\linewidth}
\centering
\includegraphics[width=1.0\linewidth]{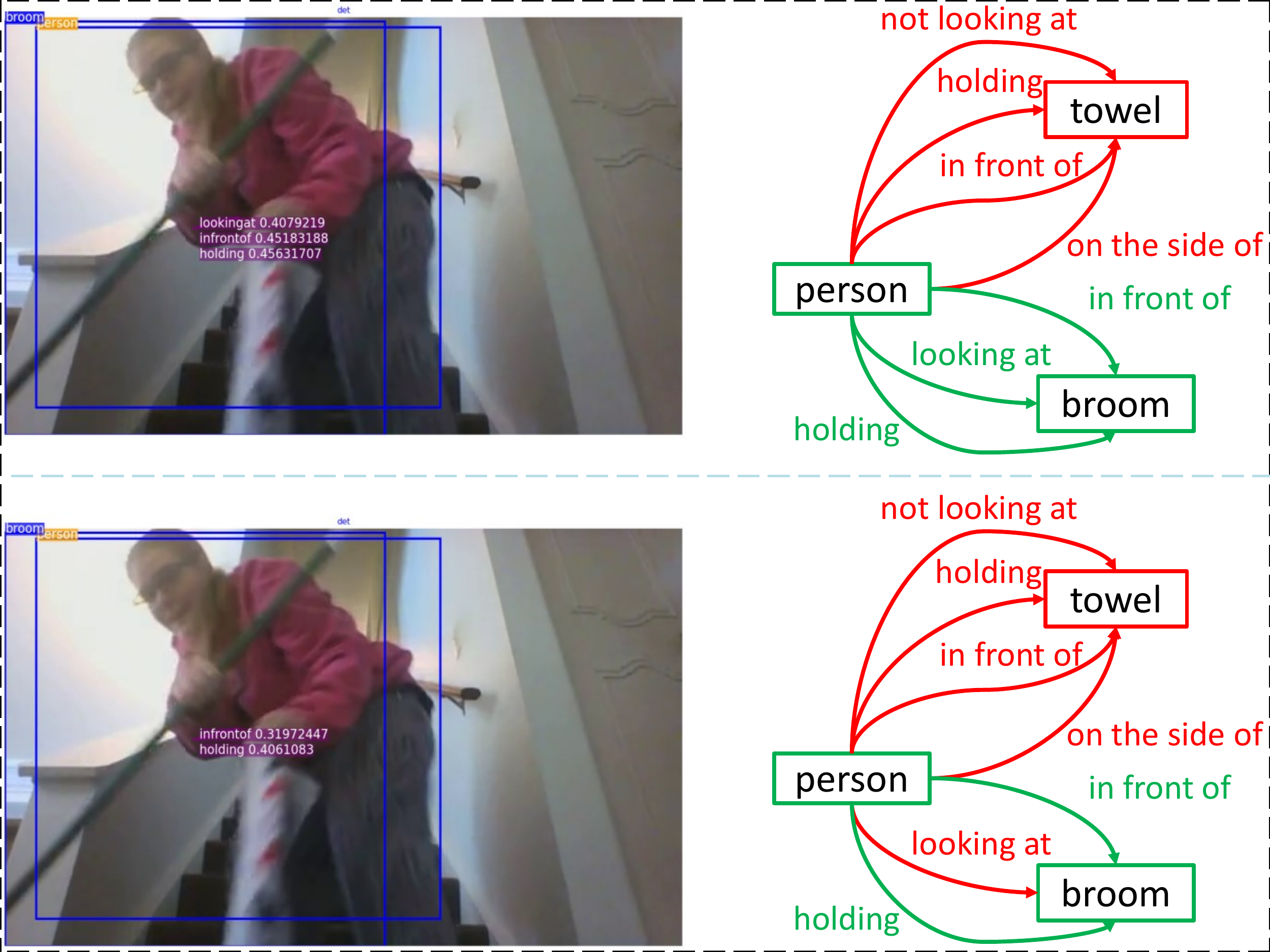}

\end{minipage}
\begin{minipage}[t]{0.47\linewidth}
\centering
\includegraphics[width=1.0\linewidth]{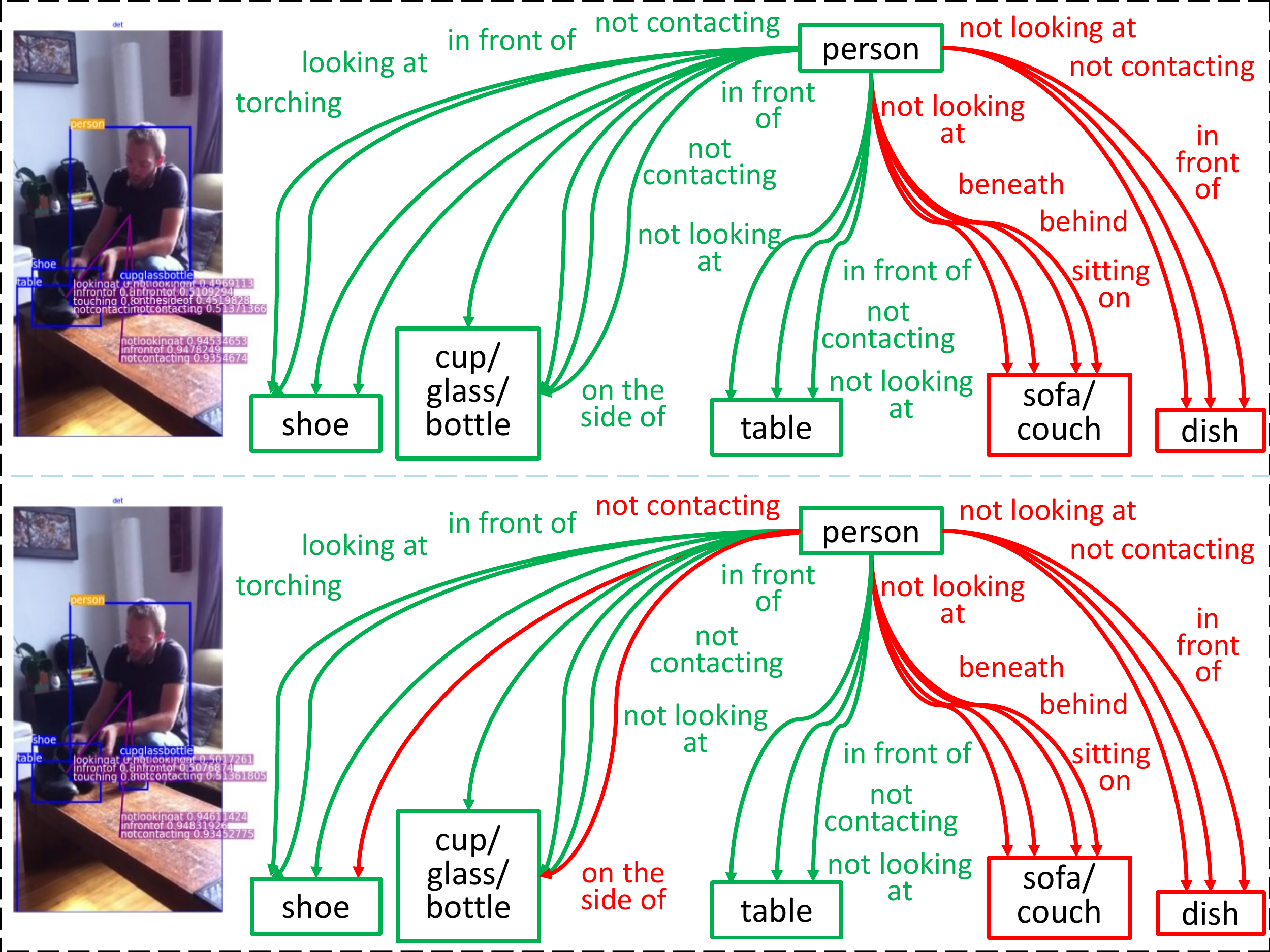}
\end{minipage}
\begin{minipage}[t]{0.47\linewidth}
\centering
\includegraphics[width=1.0\linewidth]{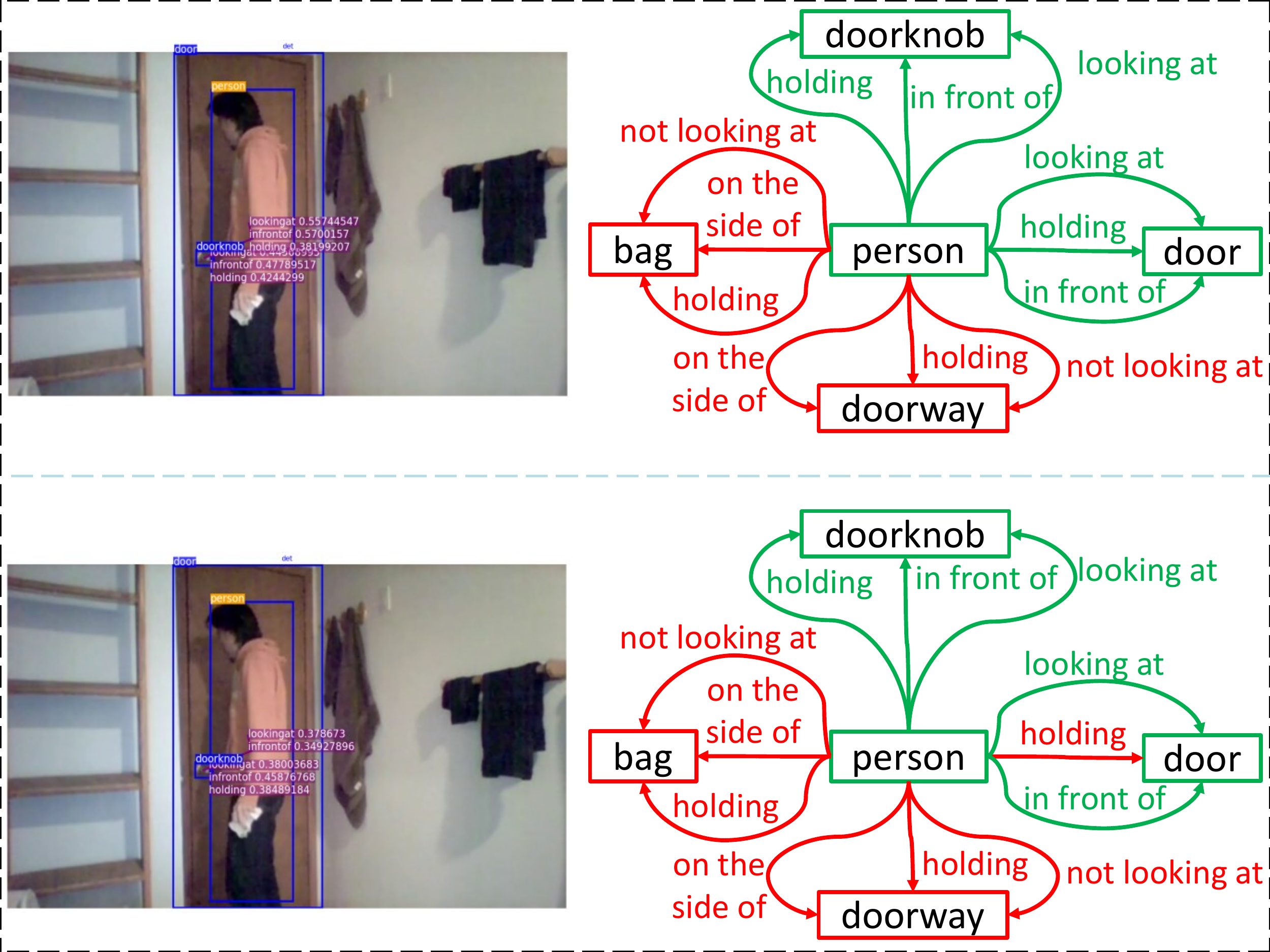}
\end{minipage}

\centering
   \caption{Qualitative comparisons between our model and RelDN~\cite{Graphical_Contrastive_Losses_for_Scene_Graph_Parsing} at Recall@20 on SGDet in AG~\cite{ag}. In each group (black dashed boxes), the top graph is our result and the bottom one is the output of RelDN~\cite{Graphical_Contrastive_Losses_for_Scene_Graph_Parsing}. In each graph, green boxes are objects which are contained in the predicted triplets and have IOU larger than 0.5 with the ground-truth boxes. Green edges are predicted relations which hit the ground-truth. Red boxes and edges are the ground-truth objects and relations which have no match with the results.}
\label{Fig:ag_comp}
\vspace{-2.5mm}
\end{figure*}

\begin{figure*}
\centering
\subfigure[Frame 1.]{
\begin{minipage}[t]{0.47\linewidth}
\centering
\includegraphics[width=1.0\linewidth]{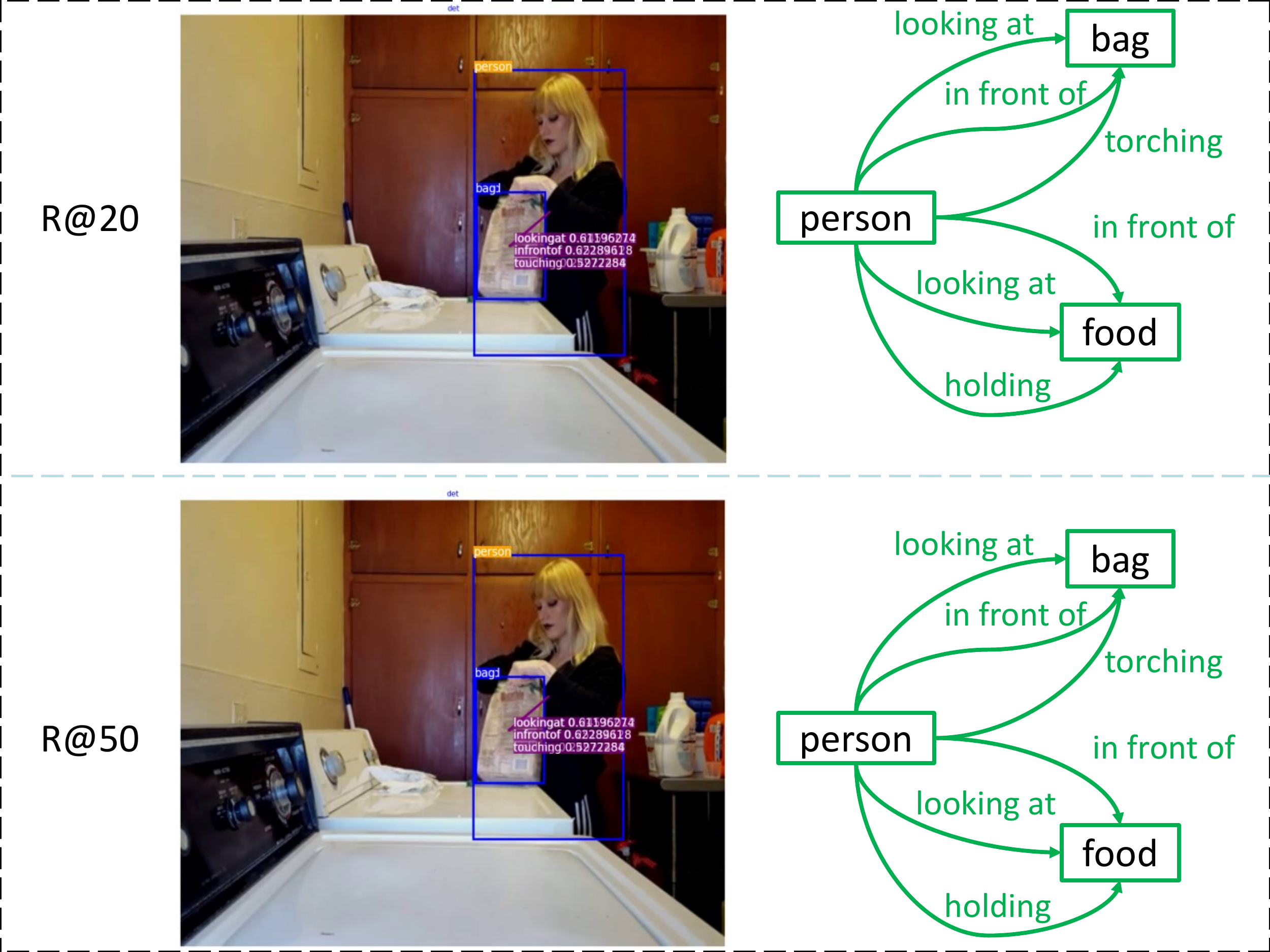}
\end{minipage}
}
\subfigure[Frame 2.]{
\begin{minipage}[t]{0.47\linewidth}
\centering
\includegraphics[width=1.0\linewidth]{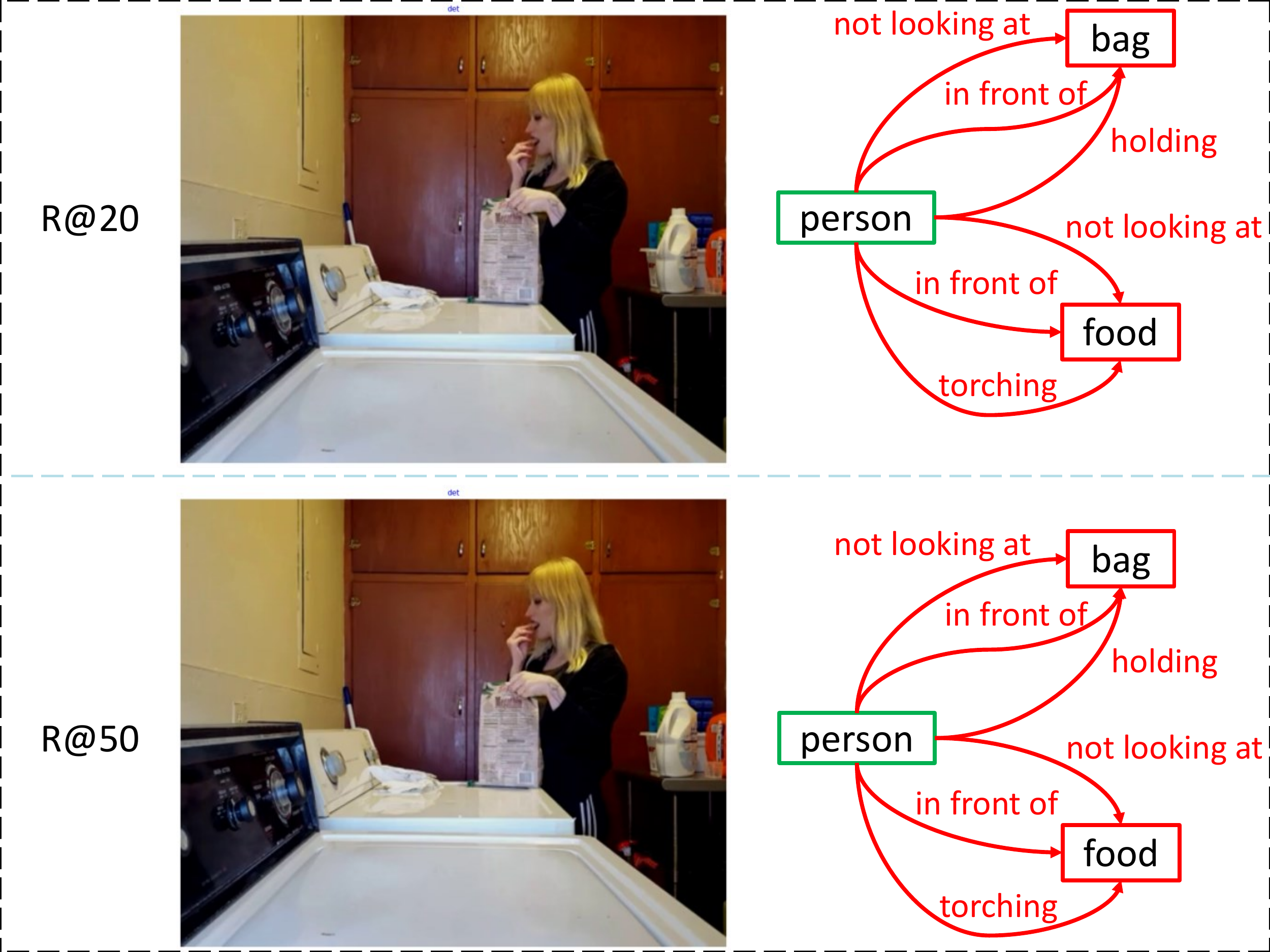}
\end{minipage}
}
\subfigure[Frame 3.]{
\begin{minipage}[t]{0.47\linewidth}
\centering
\includegraphics[width=1.0\linewidth]{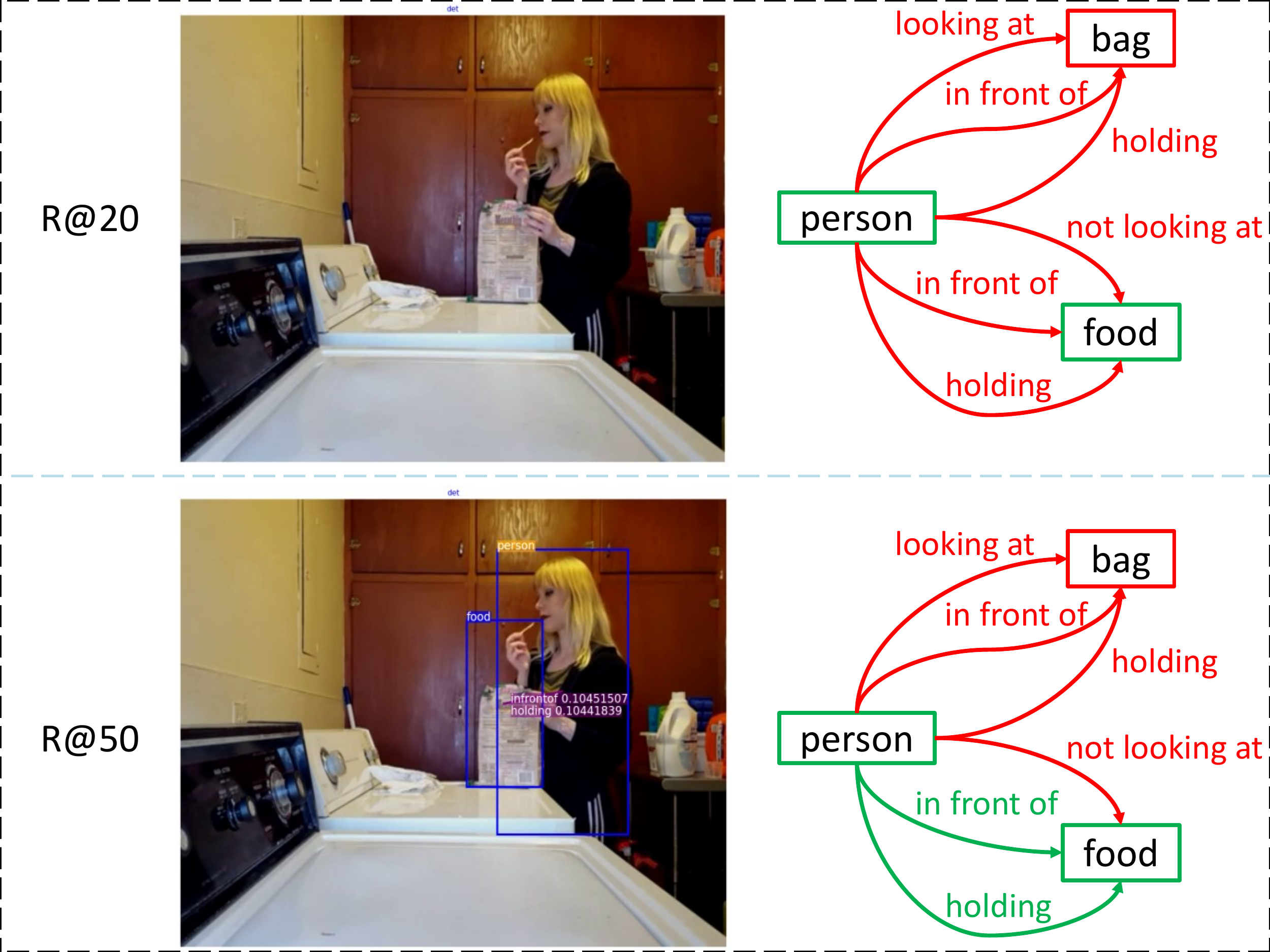}
\end{minipage}
}
\subfigure[Frame 4.]{
\begin{minipage}[t]{0.47\linewidth}
\centering
\includegraphics[width=1.0\linewidth]{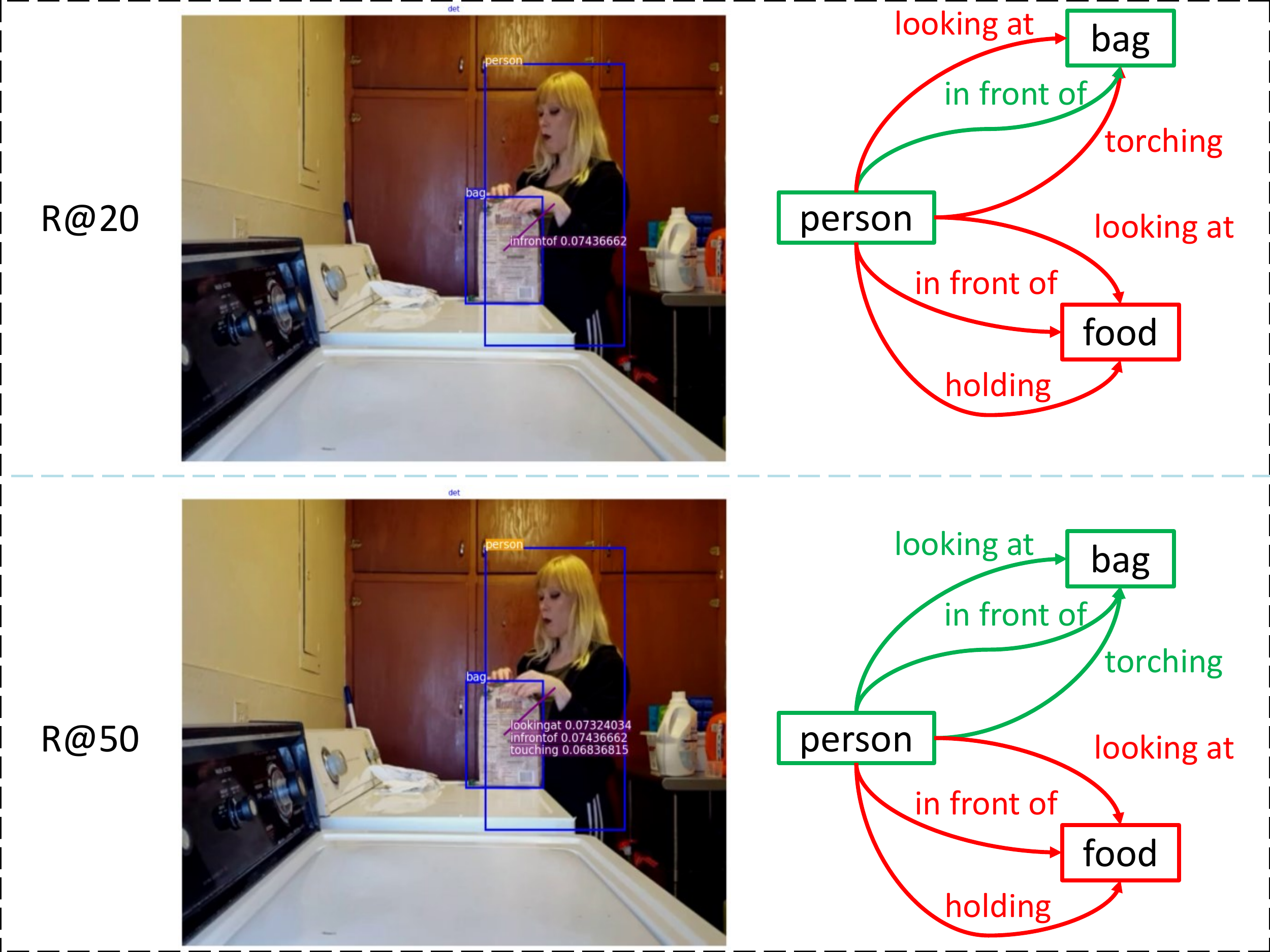}
\end{minipage}
}
\centering
   \caption{Qualitative results of our model at Recall@20 and Recall@50 on SGDet in frames sampled from a single video. In each group (black dashed boxes), the top graph is our result at Recall@20 and the bottom one is the output at Recall@50. In each graph, green boxes are objects which are contained in the predicted triplets under each metric and have IOU larger than 0.5 with the ground-truth boxes. Green edges are predicted relations which hit the ground-truth. Red boxes and edges are the ground-truth objects and relations which have no match with the results.}
\label{Fig:ag_continue}
\vspace{-2.5mm}
\end{figure*}

In Figure~\ref{Fig:ag_comp}, we provide the examples of frame-level video scene graphs generated by our model and RelDN~\cite{Graphical_Contrastive_Losses_for_Scene_Graph_Parsing} in AG dataset~\cite{ag}. In Figure~\ref{Fig:ag_continue}, we provide the examples of frame-level video scene graph generation on frames sampled from the same video clip.

{\small
\bibliographystyle{ieee_fullname}
\bibliography{egbib}
}

\end{document}